\documentclass[12pt]{article}
\usepackage{arxiv}

\usepackage[utf8]{inputenc} %
\usepackage[T1]{fontenc}    %
\usepackage{url}            %
\usepackage{booktabs}       %
\usepackage{amsfonts}       %
\usepackage{nicefrac}       %
\usepackage{microtype}      %
\usepackage{xcolor}         %
\usepackage[inline]{enumitem} %
\usepackage{natbib}
\usepackage{doi}
\usepackage{algorithm,algpseudocode}
\usepackage{amsmath}
\usepackage{amssymb}
\usepackage{mathtools}
\usepackage{amsthm}
\usepackage{mathrsfs}
\usepackage{multirow, array, threeparttable}
\usepackage[font=normalsize,skip=4pt]{caption}
\usepackage{subcaption}
\usepackage{graphicx}   %
\hypersetup{
  colorlinks,
  linkcolor={blue!60!black},
  citecolor={red!60!black},
  urlcolor={blue!80!black}
}
\usepackage[capitalize,nameinlink]{cleveref}

\makeatletter
\renewcommand{\paragraph}{%
  \@startsection{paragraph}{4}{\z@}%
  {0.25\baselineskip}  %
  {-0.25em}            %
  {\normalfont\normalsize\bfseries}%
}
\makeatother

\usepackage{amsopn} %
\usepackage{tikz} %
\usetikzlibrary{decorations.pathreplacing}
\usepackage[strings]{underscore}
\usepackage[strict=true]{csquotes}
\usepackage{siunitx} %
\DeclareMathOperator*{\argmin}{arg\,min} %

\newcommand{\glmcoef}{\boldsymbol{\theta}} %
\newcommand{\glmcoefestim}{\widehat{\glmcoef}} %
\newcommand{\residual}{\widehat{\mathbf{r}}} %
\newcommand{\ftrue}{f_{\text{true}}} %
\newcommand{\betatrue}{\beta_{\text{true}}} %

\def\approxprop{%
  \def\p{%
    \setbox0=\vbox{\hbox{$\propto$}}%
    \ht0=0.6ex \box0 }%
  \def\s{%
    \vbox{\hbox{$\sim$}}%
  }%
  \mathrel{\raisebox{0.7ex}{%
      \mbox{$\underset{\s}{\p}$}%
    }}%
}

\title{Bayesian Optimization under Uncertainty for Training a Scale Parameter in Stochastic Models}

\author{{\hspace{1mm}Akash Yadav} \\
  University of Houston\\
  \texttt{ayadav4@uh.edu} \\
  \And
  {\hspace{1mm}Ruda Zhang} \\
  University of Houston\\
  \texttt{rudaz@uh.edu} \\
}

\date{}

\renewcommand{\headeright}{ }
\renewcommand{\shorttitle}{SO-BO-scale}

\graphicspath{{img/}}

\begin{document}
    
\maketitle

\begin{abstract} %
Hyperparameter tuning is a challenging problem
especially when the system itself involves uncertainty.
Due to noisy function evaluations, optimization under uncertainty can be computationally expensive.
In this paper, we present a novel Bayesian optimization framework tailored for hyperparameter tuning under uncertainty,
with a focus on optimizing a scale- or precision-type parameter in stochastic models.
The proposed method employs a statistical surrogate for the underlying random variable,
enabling analytical evaluation of the expectation operator.
Moreover, we derive a closed-form expression for the optimizer of the random acquisition function,
which significantly reduces computational cost per iteration.
Numerical experiments demonstrate substantial gains in optimization efficiency:
about 40-fold reduction in data usage relative to a Monte Carlo–based one-dimensional optimization scheme, and an order of magnitude improvement relative to a Gaussian process-based Bayesian optimization baseline.
\end{abstract}


\section*{Practical Applications}
Accurate uncertainty quantification is essential for reliability assessment
and risk-informed engineering decisions.
Many workflows hinge on calibrating one scale/precision parameter
that controls the spread of a stochastic simulator.
Examples include tuning model-discrepancy variance in a digital twin
or setting dispersion of a stochastic model to bound prediction error.
This is an optimization-under-uncertainty problem:
each trial yields a noisy misfit that aggregates weakly correlated responses
(sum of squares or other vector norm) of the potentially high-dimensional stochastic system output,
and Monte Carlo-based searches may require thousands of runs.
This work introduces a specialized Bayesian optimization algorithm that models
this summary statistic and evaluates the expected value of the objective function analytically,
so simulations concentrate on the most informative settings.
In the case studies, the optimum was identified with about 280-290 evaluations,
using up to 40 times fewer runs than the Monte Carlo search
and 15 times fewer than standard Gaussian-process Bayesian optimization baselines.
Use this approach when simulations are costly, data are noisy,
and one global uncertainty scale must be trained.

\section{Introduction}

Optimization under uncertainty is a fundamental challenge in decision-making,
arising whenever system parameters or states are uncertain and need to be modeled as
random variables or stochastic processes. 
Such problems appear across diverse domains, including
process scheduling and planning \citep{Li2008,Verderame2010},
system operations \citep{ZhangRD2020dsp,ZhangRD2022driver},
hazard management \citep{ZhangRD2020EnvEcon} and engineering design \citep{Agarwal2004,Agarwal2004a}.
Over the years, several approaches have been developed to address problems of this nature. 

Stochastic programming \citep{Dantzig1955,birge2011introduction}
is one of the earliest formal frameworks for optimization under uncertainty,
and a detailed survey of its applications to process systems is given in \citet{Li2021}. 
Robust optimization, introduced by \citet{Bai1997} and later applied to truss topology design \citep{BenTal1997},
has since been substantially advanced \citep{BenTal2002,Bertsimas2011}, with a comprehensive review available in \citet{Beyer2007}.
Chance-constrained optimization, first formulated by \citet{Charnes1959} for temporal planning problems,
has also seen extensive development \citep{Miller1965,Nemirovski2007} and applications across finance, engineering,
and operations research.
Simulation-based optimization is an active area within stochastic optimization.
Comprehensive reviews of its developments are available in \citet{Carson1997} and \citet{Andradottir}, while simulation-based design under uncertainty is surveyed in the special issue by \citet{Li2016}.

Bayesian optimization (BO) is a widely used framework for global optimization of expensive black-box functions \citep{Jones1998},
with prominent applications in machine learning hyperparameter tuning \citep{Snoek2012,Bergstra2011} and neural architecture search \citep{White2021}.
For example, \citet{turner21a} demonstrated that BO can outperform random search for hyperparameter tuning.
A standard BO scheme fits a probabilistic surrogate model to observed data and selects new evaluation points by maximizing an acquisition criterion
that balances exploration and exploitation \citep{Frazier2018}.
Gaussian processes (GPs) are a common surrogate choice due to their flexibility and analytically tractable posteriors under Gaussian observation models \citep{Rasmussen2005},
with approximate inference typically used under non-Gaussian likelihoods.
Common acquisition functions include probability of improvement (PI), expected improvement (EI), and upper confidence bound (UCB); sampling-based policies such as Thompson sampling (TS) provide an alternative mechanism for exploration--exploitation trade-offs.
More recently, BO has been extended for optimization under uncertainty \citep{Beland2017,Oliveira2019,Yang2023}, 
with multifidelity approaches surveyed in \citet{Do2025} and \citet{ZhangRD2024mfml}. 
These developments motivate BO for structured uncertainty-aware tuning tasks beyond standard black-box settings.

In this work, we are interested in a specific problem, i.e., 
the optimization of a scale- or precision-type parameter in stochastic models.
A \textit{scale parameter} of a probability measure controls
how spread out the distribution is around its central value;
depending on context, it may be referred to as variance, dispersion, bandwidth, or temperature.
Conversely, a \textit{precision} (or \textit{concentration}) \textit{parameter} controls
how tightly the distribution clusters around the central value.
Scale and precision parameters are connected by monotone transforms (e.g., precision = 1 / variance)
and can be treated in a unified framework.
Accurate optimization of these parameters is essential for sharp, reliable uncertainty quantification.
However, because the system under study is stochastic and often complex and high-dimensional,
tuning such hyperparameters becomes a challenging problem of optimization under uncertainty.

This problem is motivated by the authors' recent work on model error characterization
using a stochastic reduced-order modeling framework \citep{yadav2025ss}.
The stochastic subspace model therein contains a hyperparameter,
denoted by $\beta$, which controls the spread of the distribution. 
The optimization approach adopted in \cite{yadav2025ss} estimated the expectation using Monte Carlo sample averages
and then optimized the estimated objective function with a one-dimensional optimization algorithm
such as golden section search and successive parabolic interpolation.
This strategy suffers from two limitations:
(i) Monte Carlo estimates of the expectation operator are noisy
and fail to exploit the global structure in the objective function,
and (ii) data inefficiency becomes severe when pointwise noise dominates the underlying function variability. 
Related stochastic reduced-order modeling approaches \citep{Soize2017srob,WangHR2019,Azzi2022}
often rely on trial-and-error with large numbers of Monte Carlo samples, which exacerbates the same inefficiency.

To overcome these limitations, we introduce a novel Bayesian optimization framework
that addresses optimization under uncertainty with significantly improved data efficiency. 
We demonstrate its effectiveness in optimizing a scale/precision parameter in stochastic models.
Our main contributions are:
\begin{enumerate}
    \item Development of a statistical surrogate for the objective function, enabling analytical evaluation of the expectation operator without Monte Carlo sampling.
    \item Derivation of a closed-form solution for the optimum of the random acquisition function, enabling efficient selection of new observation points and reducing per-iteration computational cost.
    \item Validation through numerical experiments, showing that the proposed method achieves up to a 40-fold improvement in both data usage and computational cost over the Monte Carlo search and 15-fold improvement over standard Gaussian-process Bayesian optimization baselines, making it suitable for real-time prediction tasks.

\end{enumerate}
The remainder of the paper is organized as follows. \Cref{sec:problem} introduces the optimization under uncertainty problem.
\Cref{sec:methodology} presents the proposed Bayesian optimization framework.
\Cref{sec:SROM} describes its application to optimizing a concentration parameter in stochastic models.
Numerical results are reported in \cref{sec:examples}, followed by discussion in \cref{sec:discussion} and concluding remarks in \cref{sec:conclusion}.

\section{Problem statement} \label{sec:problem}
We consider the problem of optimizing a scale- or precision-type parameter $\beta>0$
in the presence of uncertainty, formulated as:
\begin{equation} \label{eq:abstract-problem}
    \min_{\beta \in (0, \infty)} \mathbb{E}[g(s(\boldsymbol{\omega})) | \beta],
\end{equation}
where $\boldsymbol{\omega}$ is a (potentially high-dimensional) random variable,
$s(\boldsymbol{\omega}) \ge 0$ is a summary statistic and $g(\cdot)$ is a known function.
The hyperparameter $\beta$ controls the spread of the distribution of $\boldsymbol{\omega}$,
such that increasing $\beta$ increases variability in the statistic;
or inversely, it controls the concentration of the distribution of $\boldsymbol{\omega}$,
such that increasing $\beta$ reduces variability in the statistic.

The formulation in \cref{eq:abstract-problem} is intentionally kept broad:\
different problems may encounter different choices of
the function $g(\cdot)$.
In this work, we focus on the case where
$g(x) = |x - s_0|^2$, where $s_0$ is a target statistic against which $s(\boldsymbol{\omega})$ is compared.
However, the proposed methodology is not limited to this setting and can be adapted to other forms of $g$.
Under this choice, the optimization problem becomes:
\begin{equation} \label{eq:objective-abstract}
    \ftrue(\beta) := \mathbb{E}[|s(\boldsymbol{\omega}) - s_0|^2 \; | \beta],
\end{equation}
with the optimal hyperparameter
\begin{equation}
    \betatrue^* := \argmin_{\beta \in (0, \infty)} \ftrue(\beta).
\end{equation}

\section{Methodology} \label{sec:methodology}
The primary challenge in \cref{eq:objective-abstract} is that the expectation operator is typically unavailable in closed form.
While it can be approximated using Monte Carlo sample averages,
high variability in the random variable $\boldsymbol{\omega}$
demands large sample sizes for accurate approximation.
As a result, the approach requires a large number of function evaluations, which can be computationally expensive.
To address this challenge, we develop a BO framework tailored to optimization under uncertainty.
While BO methods are known to be data-efficient and well-suited for
expensive-to-evaluate objective functions,
the novelty of our approach lies in designing a BO method
that is tailored for optimization under uncertainty.
Unlike standard BO, which places a surrogate directly on the objective $f(\beta)$ and treats its evaluations as noisy black-box observations, we construct a statistical surrogate for the underlying random statistic $s(\boldsymbol{\omega})|\beta$.
This surrogate induces an analytically tractable model for $f(\beta)$, enabling closed-form evaluation of the expectation in \cref{eq:objective-abstract}.
Due to the structure of the surrogate, a further advantage of this BO approach
is that the derived surrogate for the objective function can be optimized analytically.

\subsection{Assumption}
\begin{figure}[!t]
\centering
\includegraphics[width=0.9\textwidth]{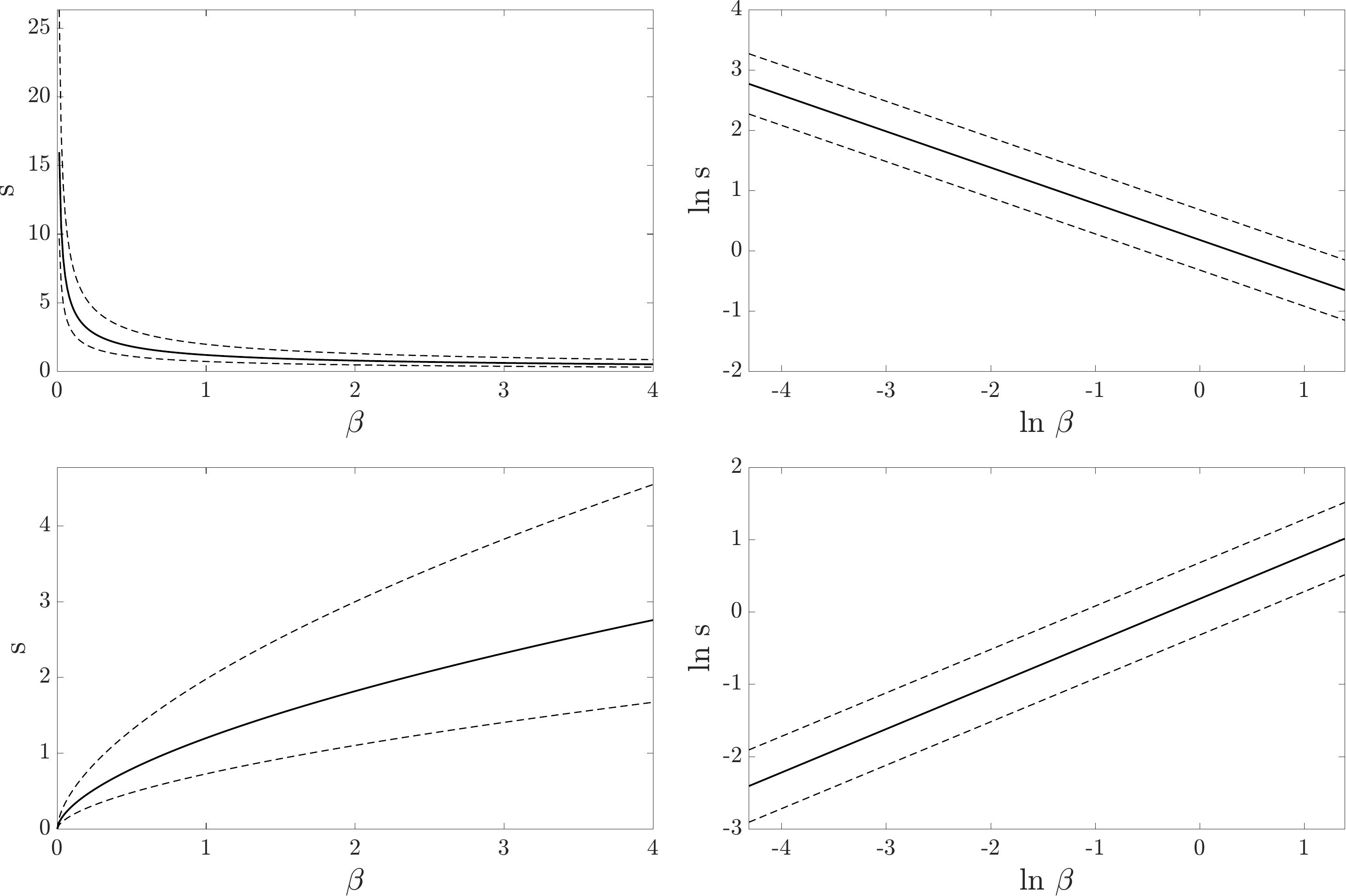}
\caption{Conditional model of $s|\beta$. The left panel shows the mean of s as a power function of $\beta$ for positive and negative exponents, with dashed curves indicating the 2.5th and 97.5th conditional percentiles. The right panel presents the corresponding log–log transformations ($\ln s$ vs. $\ln \beta$), where the relationship is linear.}
\label{fig:assumption}
\end{figure}

A key structural assumption in this work is that the expectation of the statistic scales approximately
as a power law (see \cref{fig:assumption}):
\begin{equation}\label{eq:assumption}
    \mathbb{E}[s(\boldsymbol{\omega})|\beta] \approxprop \beta^a.
\end{equation}
This assumption is consistent with many statistical settings.
A classical example is when $\boldsymbol{\omega} = (x_1, \cdots, x_n)$ is a random sample,
the scale parameter $\beta$ is the variance $\sigma^2$ of the population,
and the statistic $s(\boldsymbol{\omega}) = |\bar{x} - \mathbb{E}[x]|^2$ is the squared error of the sample mean
$\bar{x} = \frac{1}{n} \sum_{i=1}^n x_i$.
In this setting, reducing $\beta$ (i.e., lower variance or, equivalently, higher precision)
reduces the variability of the sample mean, and the statistic vanishes as $\beta$ tends to zero.
The expectation of the statistic is exactly linear in the scale parameter:
$\mathbb{E}[s(\boldsymbol{\omega})|\beta] = \frac{1}{n} \sigma^2$.
This relation changes under different parameterizations (e.g., standard deviation $\sigma$ or precision $\sigma^{-2}$),
but remains a power law with various exponents (e.g., 2 or -1).
Another example is when we take the parameter $\beta$ in the previous example to be sample size $n$ instead.
From the central limit theorems, we know that sample mean $\bar{x}$ converges to population mean $\mathbb{E}[x]$
at a polynomial rate of $-\tfrac{1}{2}$, as sample size $n$ tends to infinity.
Therefore we have $\mathbb{E}[s(\boldsymbol{\omega})|\beta] \approxprop n^{-1}$ asymptotically.
More broadly, the assumption in \cref{eq:assumption} often applies when the summary statistic $s(\boldsymbol{\omega})$ involves
averaging over weakly correlated random variables, such as through the Euclidean norm or other vector $p$-norms.
This regime is particularly important in practice and provides much of the motivation for our assumption above.

\subsection{Surrogate modeling} \label{sec:surrogate-modeling}
A central idea of the proposed method is to build a probabilistic surrogate for the statistic $s(\boldsymbol{\omega})$
conditioned on the scale or precision parameter $\beta$, exploiting its approximate power-law scaling.
Consider the classical example in the previous subsection as a prototype.
For a Gaussian random variable $x \sim N(\mu, \sigma^2)$,
we know that the sample mean $\bar{x} \sim N(\mu, \sigma^2 / n)$.
In other words, $\bar{x} = \mu + \frac{\sigma}{\sqrt{n}} z$ where $z \sim N(0, 1)$.
Therefore, we can write:
\begin{equation} \label{eq:glm-example}
    \ln |\bar{x} - \mu| = -\frac{1}{2} \ln n + \frac{1}{2} \ln \sigma^2 + \ln |z|.
\end{equation}

Based on this, we propose a conditional statistical model for $s(\boldsymbol{\omega})$ given $\beta$
in the form of a Bayesian generalized linear model (GLM):
\begin{equation} \label{eq:glm}
    \ln s(\boldsymbol{\omega}) = a \ln \beta + \ln b + \varepsilon z, \quad z \sim N(0, 1),
\end{equation}
where $a$ is the exponent of the power law, $b$ is a scaling parameter,
and $\varepsilon$ captures variability around the power-law trend. 
Here we adopt an independent Gaussian noise $z \sim N(0,1)$ for analytical tractability;
alternative noise models are discussed in \cref{sec:discussion}.
For simplicity, we use the standard noninformative prior for the GLM parameters $(a, \ln b, \varepsilon)$.
Denote coefficients $\glmcoef = (a, \ln b)$,
the standard noninformative prior is uniform on $(\glmcoef, \ln \varepsilon^2)$
and has an improper distribution: $p(\glmcoef, \varepsilon^2) \propto \varepsilon^{-2}$.
The distribution is improper because it has an infinite integral over its domain
$\mathbb{R}^2 \times (0, \infty)$.

Notice that, with the surrogate model described above,
the random variable that drives the uncertainty in $s$
reduces from the potentially high-dimensional $\boldsymbol{\omega}$
to a four-dimensional one: $(\glmcoef, \varepsilon^2, z)$.
Such a simplification in uncertainty representation aims to capture the most salient features
of the conditional distribution $s|\beta$---as summarized by its leading moments---rather than its finer distributional details.
This parsimonious parameterization supports sample-efficient estimation during the optimization process.

\begin{figure}[!t]
\centering
\includegraphics[width=0.75\textwidth]{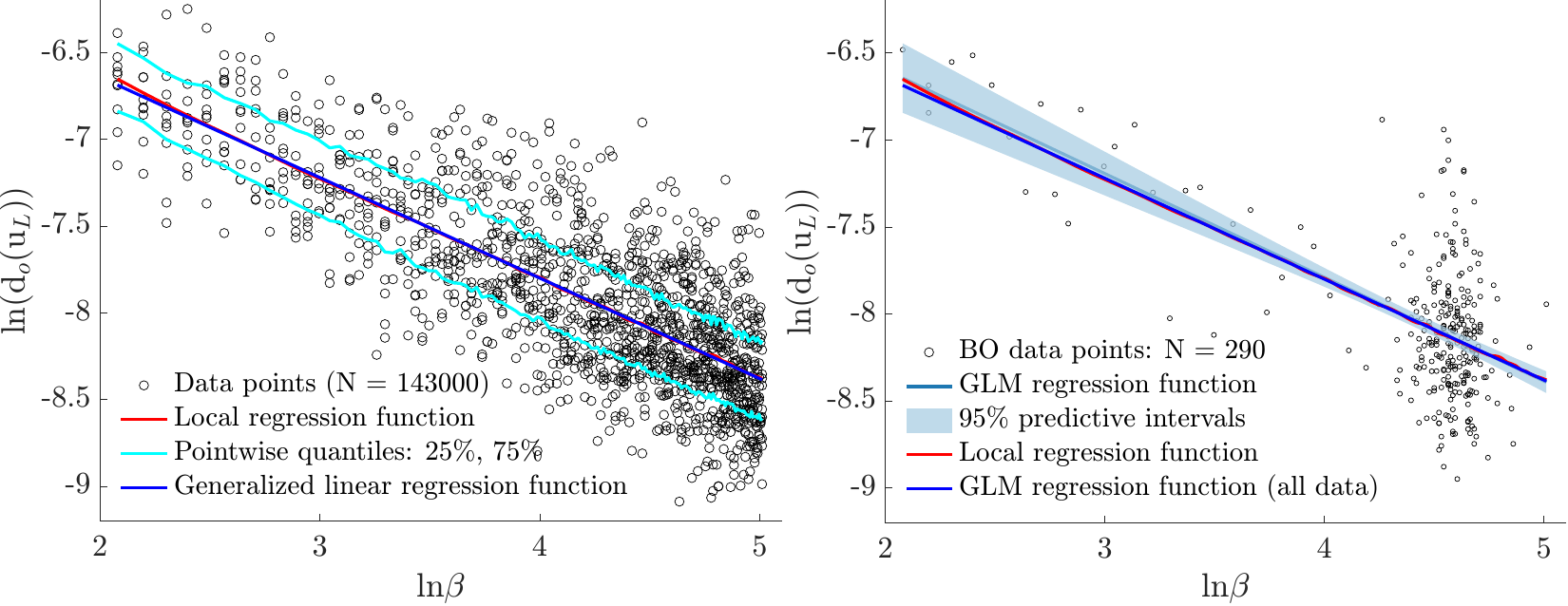}
\caption{Static problem: raw data and surrogate model fit. (Left) Raw data is highly noisy with a downward linear trend.
Plotted points are 1\% of the data, uniformly subsampled.
(Right) Bayesian GLM constructed during Bayesian optimization captures the GLM regression function accurately with quantified error,
using much fewer data points.}
\label{fig:glm-Bayes}
\end{figure}

\textit{Inference.}
For the GLM in \cref{eq:glm}, denote input $\mathbf{x} = (\ln \beta, 1)$ and output $y = \ln s(\boldsymbol{\omega})$.
Given a dataset $\mathcal{D} = \{(\mathbf{x}^i, y^i)\}_{i=1}^n$
with an input sample $\mathbf{X} = [\mathbf{x}^1 \cdots \mathbf{x}^n]^\intercal \in \mathbb{R}^{n \times k}$
and output observations $\mathbf{y} = (y^1, \cdots, y^n)$,
the classical estimates of the GLM parameters $(\glmcoef, \varepsilon^2)$ are:
\begin{equation} \label{eq:glm-estimate}
    \glmcoefestim = \mathbf{X}^\dagger \mathbf{y}, \qquad
    s^2 = \frac{1}{n-k} \residual^\intercal \, \residual,
\end{equation}
where $\dagger$ denotes the Moore--Penrose inverse and $\residual = \mathbf{y} - \mathbf{X} \glmcoefestim$ is the residual.
If $\mathbf{X}$ has full column rank, then $\mathbf{X}^\dagger = (\mathbf{X}^\intercal \mathbf{X})^{-1} \mathbf{X}^\intercal$.
With the standard noninformative prior, the posterior distribution of the GLM parameters can be written as:
\begin{equation} \label{eq:glm-posterior}
    \varepsilon^2 | \mathcal{D} \sim \text{Inv-}\chi^2(n-k, s^2), \qquad
    \glmcoef | \varepsilon^2, \mathcal{D} \sim N(\glmcoefestim, \varepsilon^2 \mathbf{V}_{\glmcoef}),
\end{equation}
where $\mathbf{V}_{\glmcoef} = (\mathbf{X}^\intercal \mathbf{X})^{-1}$.
The scaled inverse chi-squared distribution $\text{Inv-}\chi^2(\nu, s^2)$
is the distribution of $\nu s^2 / x$ where $x \sim \chi^2(\nu)$
is a chi-squared random variable with $\nu$ degrees of freedom.
For any input sample
$\widetilde{\mathbf{X}} = [\widetilde{\mathbf{x}}^1 \cdots \widetilde{\mathbf{x}}^m]^\intercal \in \mathbb{R}^{m \times k}$,
the Bayesian GLM gives a prediction $\widetilde{\mathbf{y}} = (\widetilde{y}^1, \cdots, \widetilde{y}^m)$
with the joint distribution:
\begin{equation}
    \widetilde{\mathbf{y}}|\mathcal{D} \sim t_m(\widetilde{\mathbf{X}} \glmcoefestim, s^2 (\mathbf{I}_m + \widetilde{\mathbf{X}} \mathbf{V}_{\glmcoef} \widetilde{\mathbf{X}}^\intercal), n-k),
\end{equation}
where $t_m(\boldsymbol{\mu}, \boldsymbol{\Sigma}, \nu)$ is the $m$-dimensional $t$-distribution
with mean $\boldsymbol{\mu}$, covariance $\boldsymbol{\Sigma}$, and degrees of freedom $\nu$.
\Cref{fig:glm-Bayes} shows the GLM fit using a large number of data points
and a Bayesian GLM fit during the Bayesian optimization process.
We observe that both the GLM and the Bayesian GLM capture the downward trend in the data. 
The data used here comes from the static problem described in \cref{sec:static}.

The Bayesian GLM also induces a surrogate model on the objective function.
From \cref{eq:glm}, the conditional model can be written as:
\begin{equation} \label{eq:glm-do}
    s(\boldsymbol{\omega})|\beta = b \beta^a \zeta, \qquad \zeta = \exp(\varepsilon z) \sim \ln N(0, \varepsilon^2).
\end{equation}
Here $\zeta$ has a log-normal distribution $\ln N(0, \varepsilon^2)$
because $\ln \zeta = \varepsilon z \sim N(0, \varepsilon^2)$,
with mean $\mathbb{E}[\zeta] = \exp(\tfrac{1}{2} \varepsilon^2)$
and variance $\mathbb{V}[\zeta] = \exp(2 \varepsilon^2) - \exp(\varepsilon^2)$.
Plugging the conditional model into \cref{eq:objective-abstract},
we obtain a surrogate model of the objective function:
\begin{align} \label{eq:f-surrogate}
    f(\beta) &= \mathbb{E}[|b \beta^a \zeta - s_0|^2]
    = \mathbb{V}[b \beta^a \zeta] + (\mathbb{E}[b \beta^a \zeta] - s_0)^2
    = b^2 \beta^{2a} \mathbb{V}[\zeta] + (b \beta^a \mathbb{E}[\zeta] - s_0)^2 \nonumber \\
    &= b^2 \beta^{2a} \left[\exp(2 \varepsilon^2) - \exp(\varepsilon^2)\right]
    + \left[b \beta^a \exp\left(\tfrac{1}{2} \varepsilon^2\right) - s_0\right]^2.
\end{align}
Here, the expectation is taken with respect to $z$.
Along with the posterior distribution of the GLM parameters in \cref{eq:glm-posterior},
this gives a probabilistic surrogate of the objective function.
We may also plug in the classical GLM estimate in \cref{eq:glm-estimate}
to obtain a deterministic surrogate of the objective function.

\textit{Discussion.}
Typical BO methods construct a probabilistic surrogate for the objective function directly,
and most of the time the surrogate is a generic Gaussian process \citep{Do2024photonics}.
Our proposed approach exploits the analytical form of the objective function,
and builds a probabilistic surrogate for a component random variable instead.
The Bayesian GLM for the random variable induces a parametric non-Gaussian random process
as a probabilistic surrogate for the objective function.

\subsection{Acquisition function}

Once a probabilistic surrogate is constructed, given the available data,
a BO algorithm selects the next observation location, evaluates the objective function,
and then updates the surrogate with the new data point.
The process iterates until the optimization is terminated.
At each iteration, the new observation location is identified as the global optimum of an acquisition function,
which is derived from the probabilistic surrogate.

In this work, the probabilistic surrogate $f(\beta)$ in \cref{eq:f-surrogate}
is a complex transform of the GLM parameters $(\glmcoef, \varepsilon^2)$,
so it is difficult to conduct inference analytically.
This creates a challenge in using many common acquisition functions for BO,
which requires computing the mean, variance, or other expectations of the surrogate.
But since it is straightforward to sample $f(\beta)$, we choose \textit{Thompson sampling} as the acquisition strategy:
a posterior sample of the objective function is used as the acquisition function,
whose global optimum is used as the next observation location.
In our case, a posterior sample of the GLM parameters can be sampled via \cref{eq:glm-posterior},
which is then plugged into \cref{eq:f-surrogate} to obtain a posterior sample of the objective function.

\textit{Discussion.}
For BO with a continuous design space, Thompson sampling is implemented in various ways,
including pointwise joint sampling and spectral sampling.
\citet{Do2025edu} gives a detailed presentation of various approaches to sampling from Gaussian process posterior.
Pointwise joint sampling creates exact samples on a finite set of design points,
by sampling from the joint distribution of the objective function values at those points.
This method becomes prohibitively costly as the number of sample points increases.

Spectral sampling creates approximate samples throughout the entire design space,
based on a representation of the objective function.
The representation could be a linear combination of basis functions,
such as the kernel eigenfunctions or the Fourier basis.
Such a representation is usually an infinite series,
and in practice, only the coefficients of the leading terms are sampled,
which defines an approximate sample of the objective function.
In comparison, our proposed approach creates exact function samples,
thanks to the parametric structure of the surrogate model.

Posterior samples can be used for the Monte Carlo approximation of many acquisition functions,
bypassing the need for analytic inference.
Sample average approximation also provides a way to control the exploration--exploitation tradeoff
in Thompson sampling \citep{Do2024egts}.
Here, we stick with the classic Thompson sampling for its simplicity and analytical advantages, as detailed below.

\subsection{Global optimization}

In BO methods, the global optimization of acquisition functions typically relies on numerical optimization algorithms.
Specialized algorithms are often needed for accurate and scalable optimization,
especially for Thompson sampling \citep{Adebiyi2024bdu,Adebiyi2025tsroots}.
But for the BO method we are proposing, the acquisition functions are the posterior sample function given in \cref{eq:f-surrogate},
whose global minimum has an analytical solution.

\begin{figure}[!t]
\centering
\includegraphics[width=0.65\textwidth]{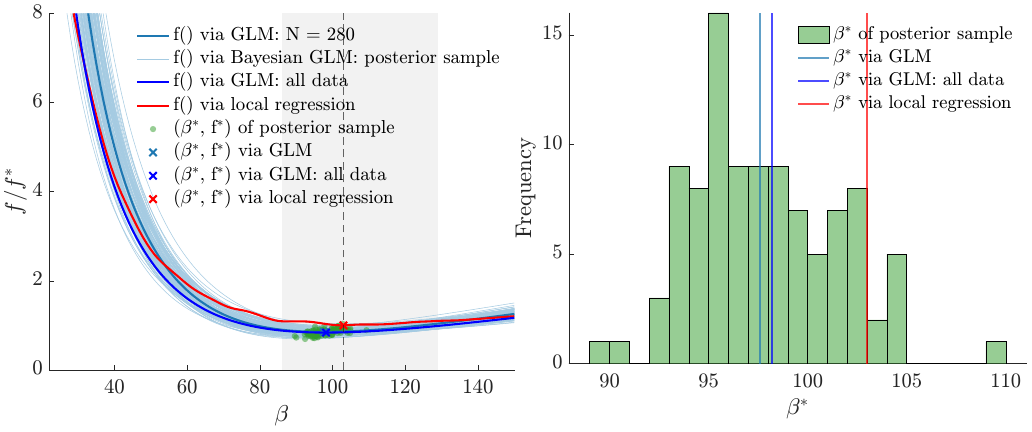}
\caption{Static problem: optimization results under different approaches. (Left) Posterior distribution of $(\beta^*, f^*)$ well captures the true statistics.
(Right) Histogram of $\beta^*$ posterior samples.}
\label{fig:f-glm-TS}
\end{figure}

The derivative of the posterior sample function can be written as:
\begin{equation*}
    f'(\beta) = b^2 \left[\exp(2 \varepsilon^2) - \exp(\varepsilon^2)\right] (2a \beta^{2a-1})
    + 2 \left[b \exp\left(\tfrac{1}{2} \varepsilon^2\right) \beta^a - s_0\right]
    [b \exp\left(\tfrac{1}{2} \varepsilon^2\right) (a \beta^{a-1})].
\end{equation*}
Let $C_1 = b^2 \left[\exp(2 \varepsilon^2) - \exp(\varepsilon^2)\right]$
and $C_2 = b \exp\left(\tfrac{1}{2} \varepsilon^2\right)$,
then we have:
\begin{equation*}
    f'(\beta) = 2a \beta^{a-1} [C_1 \beta^a + (C_2 \beta^a - s_0) C_2].
\end{equation*}
We see that the critical point condition $f'(\beta_0) = 0$ is equivalent to $\beta_0^a = s_0 C_2 / (C_1 + C_2^2)$.
Since there is only one critical point for $\beta \in (0, \infty)$, it is the only internal extremum of $f(\beta)$.
Further examination shows that $f'(\beta) < 0$ for $\beta < \beta_0$ and $f'(\beta) > 0$ for $\beta > \beta_0$,
which means that $\beta_0$ is the global minimum of $f(\beta)$. Therefore, we can write:
\begin{equation}\label{eq:beta-analytical}
    \beta^* := \argmin_{\beta \in (0, \infty)} f(\beta) = \left(\frac{s_0}{b \exp(\tfrac{3}{2} \varepsilon^2)}\right)^{1/a}.
\end{equation}
Given a posterior sample of $(a, b, \varepsilon^2)|\mathcal{D}$ via \cref{eq:glm-posterior},
the above formula gives a posterior sample of $\beta^*|\mathcal{D}$,
which is our current probabilistic estimate of the optimal hyperparameter $\betatrue$.
Plugging in the classical GLM estimate in \cref{eq:glm-estimate} gives a deterministic estimate of $\betatrue$.
\Cref{fig:f-glm-TS} shows some posterior samples of $f(\beta)$ at an iteration during the BO process,
as well as the corresponding Thompson samples $(\beta^*, f^*)$, where $f^* := f(\beta^*)$.

\subsection{Observation mode}

While optimization is fundamentally a sequence of decisions,
parallelization can sometimes be exploited to accelerate data acquisition.
For example, if the objective function is approximated via Monte Carlo sampling,
the sampling process can be parallelized trivially.
To enable parallelization in the proposed BO approach, we use a synchronous batch policy:
at each iteration, a batch of observation locations are designed and observations are obtained,
before designing the next batch of observation locations.

\begin{figure}[!t]
\centering
\includegraphics[width=0.75\textwidth]{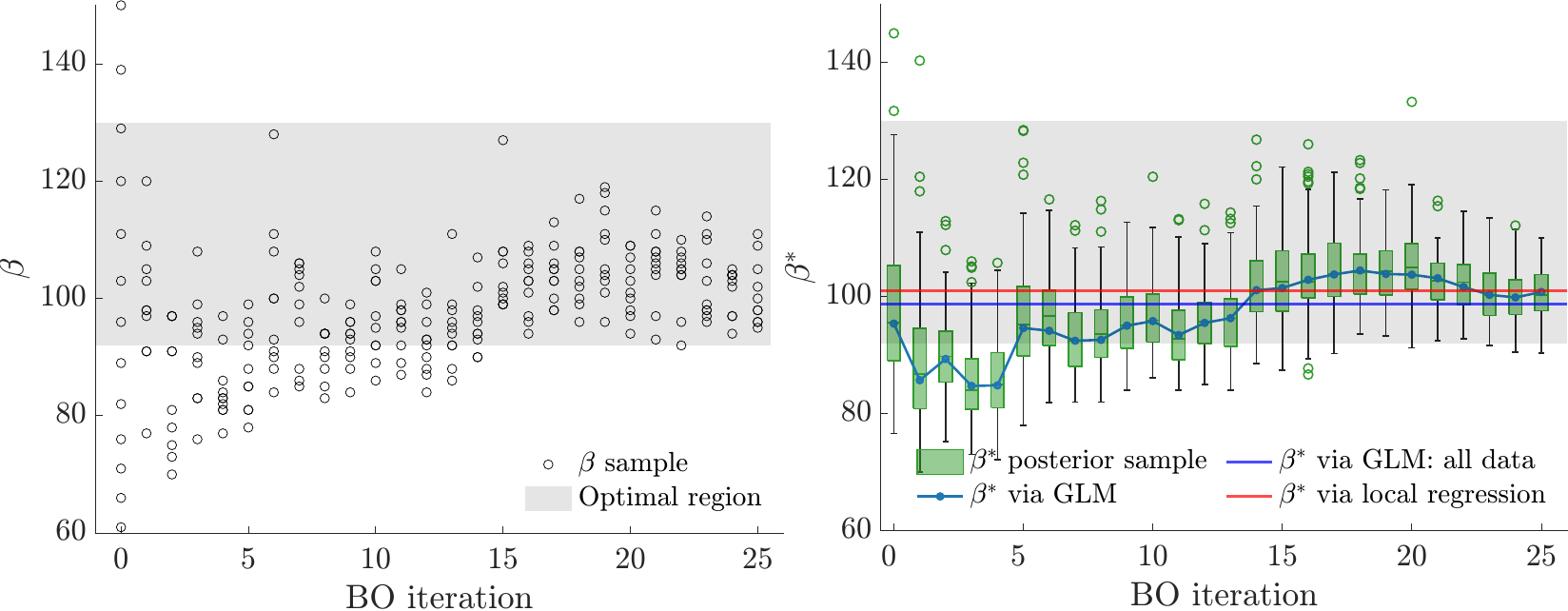}
\caption{Static problem: progression of Bayesian optimization. (Left) Sample points during BO iterations.
(Right) Evolution of $\beta^*$ posterior distribution and GLM estimate.}
\label{fig:series-BO}
\end{figure}

For Thompson sampling, it is straightforward to design a batch of observation locations:
simply generate multiple samples of $\beta^*|\mathcal{D}$.
\Cref{fig:series-BO} shows the sequence of sample points during BO iterations,
where after an initial sample of size 40, uniform in $\ln \beta$, we use a batch size of 10 for Thompson sampling.
The figure also shows the evolution of the posterior distribution $\beta^*|\mathcal{D}$ and the GLM estimate.

\begin{figure}[!tbp]
    \centering
    \includegraphics[width=0.55\textwidth]{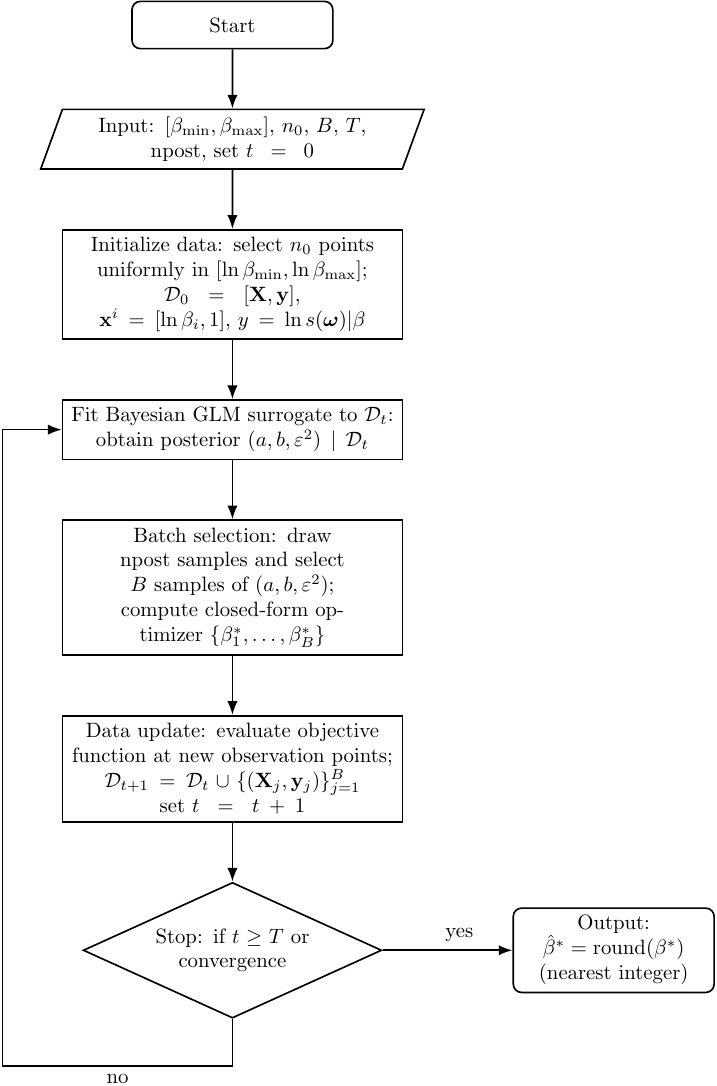}
    \caption{Flowchart of the proposed method.}
    \label{fig:flowchart}
\end{figure}

For clarity, the overall workflow of the proposed methodology is summarized in the flowchart shown in \cref{fig:flowchart}.

\section{Application: Concentration parameter optimization in stochastic models} \label{sec:SROM}
We now apply the proposed framework to optimize a concentration parameter in stochastic models.

Let $F^o(\mathbf{a})$ denotes a reference model with input $\mathbf{a}$.
Let $F_L(\mathbf{a}; \mathscr{V})$ be the low-fidelity, reduced-order model (ROM)
on a subspace $\mathscr{V}$,
and let $\mathscr{W}$ be a stochastic subspace with a concentration parameter $\beta$.
To characterize the predictive error of the reference model using a stochastic reduced-order model (SROM)
$F_L(\mathbf{a}; \mathscr{W})$,
we train the concentration parameter
so that the experimental validation data $\mathbf{u}_E$ behaves like a typical SROM prediction.
Specifically, denote the $L^2$ distance of an output to a reference prediction $\mathbf{u}^o$ as:
\begin{equation}
    d_o(\mathbf{u}) = \|\mathbf{u} - \mathbf{u}^o\|_{L^2}.
\end{equation}
The functional $d_o$ summarizes the SROM prediction
$\mathbf{u}_L = F_L(\mathbf{a}; \mathscr{W})$, a random process,
into a random variable $d_o(\mathbf{u}_L)$ which characterizes the variability of $\mathbf{u}_L$.
The objective function for the hyperparameter $\beta$
is then defined as the mean squared error of $d_o(\mathbf{u}_L)$ from $d_o(\mathbf{u}_E)$, that is:
\begin{equation} \label{eq:objective}
    \ftrue(\beta) := \mathbb{E}[|d_o(\mathbf{u}_L) - d_o(\mathbf{u}_E)|^2 \; | \beta].
\end{equation}
This is a direct instance of the abstract problem in \cref{eq:abstract-problem,eq:objective-abstract}, where random variable $\omega = \mathscr{W}$, summary statistic $s(\omega) = \|F_L(\mathbf{a}; \omega) - \mathbf{u}^o\|_{L^2}$,
and function $g(x) = |x - d_o(\mathbf{u}_E)|^2$.
The optimal hyperparameter is defined as:
\begin{equation}
    \betatrue^* := \argmin_{\beta \in [k, \infty)} \ftrue(\beta).
\end{equation}
This formulation casts hyperparameter tuning in SROMs as an optimization under uncertainty problem, 
making it directly amenable to the Bayesian optimization framework developed in \cref{sec:methodology}.
While the methodology treats $\beta$ as continuous in this particular application, $\beta$ corresponds to the number of samples and is therefore implemented as an integer (we evaluate at the nearest integer to the continuous optimum).
Furthermore, since at least $k$ samples are required to construct a reduced subspace, 
the lower bound of $\beta$ is set to the ROM dimension ($k$).
The proposed framework is implemented to tune
the SROM concentration parameter $\beta$. Given bounds $[\beta_{\min},\beta_{\max}]$, an initial
design size $n_0$, batch size $B$, and a maximum of $T$ iterations, we first generate $n_0$ initial values of
$\beta$ by sampling uniformly in log scale over $[\ln\beta_{\min},\ln\beta_{\max}]$ and evaluate the SROM at
these settings to form the initial dataset $\mathcal{D}_0=[\mathbf{X},\mathbf{y}]$, where
$\mathbf{x}^i=[\ln\beta_i,\,1]$ for $i=1,\ldots,n_0$ and $y=\ln d_o(\mathbf{u}_L)$. At iteration $t$, we fit
the Bayesian GLM surrogate in \cref{eq:glm} to $\mathcal{D}_t$ to obtain the posterior
$p(a,b,\varepsilon^2\mid \mathcal{D}_t)$. Next, we draw $B$ posterior samples of $(a,b,\varepsilon^2)$ using
\cref{eq:glm-posterior}; for each sample, we compute the associated closed-form optimizer $\beta^*$ via
\cref{eq:beta-analytical}, yielding a batch of candidates $\{\beta_1^*,\ldots,\beta_B^*\}$. We then evaluate
the SROM at these batch points and augment the dataset as
$\mathcal{D}_{t+1}=\mathcal{D}_t\cup\{(\mathbf{X}_j,\mathbf{y}_j)\}_{j=1}^B$. This
fit--sample--optimize--evaluate cycle is repeated until the stopping criterion is met (e.g., exhausting the
evaluation budget or observing convergence of $\beta^*$), after which we report the final estimate
$\hat{\beta}^*$ as the nearest-integer version of the selected $\beta^*$.

\section{Numerical example} \label{sec:examples}

This section evaluates the proposed method against three baselines: (i) one-dimensional optimization with Monte Carlo approximation of the expectation, (ii) GP-based surrogate with heteroscedastic noise model and Thompson sampling acquisition function guided by BO, and (iii) GP-based surrogate with Noisy Expected Improvement (qNEI) acquisition function guided by BO.
We consider two engineering examples to assess the accuracy and computational efficiency of the proposed method.
All the data and code for these examples are available at \href{https://github.com/UQUH/SO-BO-scale}{https://github.com/UQUH/SO-BO-scale}.

\subsection{Static problem} \label{sec:static}

\begin{figure}[!t]
    \centering
    \includegraphics[width=0.6\textwidth]{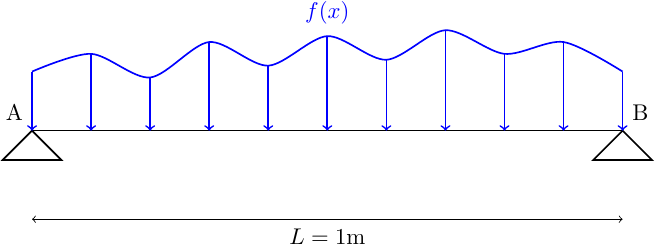}
    \caption{One-dimensional fixed-fixed static system.}
    \label{fig:diagram-static}
\end{figure}

We consider a static problem where the experimental displacement follows the model:
\begin{equation}
    \mathbf{K} \mathbf{x}_{\text{E}} = \mathbf{f}_{\text{E}},
\end{equation}
with 1,000 uniformly spaced degrees of freedom (DoFs) and boundary conditions $\mathbf{B}^\intercal \mathbf{x}_{\text{E}} = 0$
where $\mathbf{B} = [\mathbf{e}_1 \, \mathbf{e}_n]$. 
The experimental force vector is defined as: $\mathbf{f_{\text{E}}} = \frac{1}{0.261466}(0.1\phi_2 + 0.4\phi_5 + 0.6\phi_8 + 2.5(\phi_{31} + \phi_{32}) - 0.015\phi_1)$, where $\phi_i$ is the $i$-th eigenvector of the stiffness matrix.
The stiffness matrix %
is formulated as $\mathbf{K} = \mathbf{\Phi} \mathbf{\Lambda} \mathbf{\Phi}^{\intercal}$, where $\mathbf{\Lambda}$ = diag$(\lambda_i)$, $\lambda_i = 4\pi^2i^2, i = 1, \cdots, n$.
The matrix $\boldsymbol{\Phi}$ is computed by the QR decomposition of matrix $\mathbf{P}$, where $\mathbf{P} = \sin \left(k\pi \frac{j-1}{n-1}\right), j = 1, \cdots, n$ and $k = 1, \cdots, n$. 
This setup can be interpreted as a generic one-dimensional structural system with fixed–fixed boundary conditions (see \cref{fig:diagram-static}).
In practice, the experimental model is inaccessible and is approximated by a high-dimensional model (HDM), which is defined as:
\begin{equation}
    \mathbf{K} \mathbf{x}_{\text{H}} = \mathbf{f}_{\text{H}},
\end{equation}
with $\mathbf{f_{\text{H}}} = \frac{1}{0.27702}(0.1\phi_2 + 0.4\phi_5 + 0.6\phi_8 + 2.5(\phi_{29} + \phi_{30} + \phi_{31}))$ and subject to the same boundary conditions.
The experimental model and HDM model share the same system operator $\mathbf{K}$ and boundary conditions, but differ in the forces, which represent a load/input mismatch.
Because $\mathbf{f}_{\text{H}} \neq \mathbf{f}_{\text{E}}$, the HDM displacement $\mathbf{x}_{\text{H}}$ deviates from $\mathbf{x}_{\text{E}}$, giving rise to model error.

To efficiently characterize/correct this discrepancy, we employ the SROM framework.
Firstly, a deterministic ROM is constructed by projecting the HDM onto an eight-dimensional reduced subspace spanned by the leading eight eigenvectors of $\mathbf{K}$.
The ROM is defined as:
\begin{equation}
  \mathbf{x}_{\text{R}} = \mathbf{V}\mathbf{q},
  \quad \mathbf{V}^{\intercal} \mathbf{K} \mathbf{V} \mathbf{q} = \mathbf{V}^{\intercal}\mathbf{f}_{\text{H}},
\end{equation}
The SROM extends this construction by replacing the deterministic basis $\mathbf{V}$ with a stochastic basis $\mathbf{W}$, sampled from the stochastic subspace model of \citet{yadav2025ss}.
The stochastic subspace model has a hyperparameter, $\beta$, that controls the distribution's spread.
The goal is to estimate $\beta$
so that the experimental validation data $\mathbf{u}_E$ behaves like a typical SROM prediction, as formalized by the objective function in \cref{eq:objective}. For this example, $d_o(\mathbf{u}_E)$ is the $L^2$ distance between experimental and ROM displacement, 
and $d_o(\mathbf{u}_L)$ is the $L^2$ distance between ROM and SROM displacement.
The hyperparameter $\beta$ is estimated by following the methodology outlined in \cref{sec:methodology,sec:SROM}.
The results are compared with those obtained from a one-dimensional optimization algorithm with Monte Carlo sampling and with two GP-based surrogate baselines.

\Cref{fig:glm-Bayes} compares the Bayesian GLM fit obtained using 290 BO samples to a GLM fit from the full dataset of 143,000 samples.
Both capture the expected downward scaling trend, but the Bayesian GLM achieves this with dramatically fewer evaluations.
Compared with 12,000 Monte Carlo evaluations for a one-dimensional optimization scheme and 4,400 evaluations for the GP-based BO baseline, the Bayesian GLM requires only 290 samples to identify the optimum (see \Cref{tab:all-examples}). 
This translates into a 41-fold reduction in data usage compared to a one-dimensional optimization scheme and around 15-fold reduction compared to GP-based baselines.
These gains arise from two factors: (i) the GLM surrogate captures the dominant log--log linear (power-law) scaling with respect to $\beta$, and (ii) BO concentrates evaluations near the optimum.

\begin{table}[!t]
\centering
\caption{Comparison of data points, Computational time, absolute accuracy, and standard deviation of posterior samples for different methods across three examples. Ratios in parentheses are relative to Bayesian GLM.}
\label{tab:all-examples}
\setlength{\tabcolsep}{5pt}
\begin{tabular}{llrrrr}
\toprule
Example & Method & Data points & Time (s) & $\mid\beta^* - \beta^*_{LR}\mid$ & $\sigma(\beta^*)$ \\
\midrule

\multirow{4}{*}{Static}
& Bayesian GLM   & \textbf{290}     & \textbf{0.16}   & \textbf{0.18} & \textbf{3.68} \\
& GP-BO (TS)    & 4,400 $(15.1\times)$ & 3.4 $(21.2\times)$ & 5.6  & 4.5 $(1.2\times)$ \\
& GP-BO (qNEI)    & 5,100 $(17.6\times)$ & 3.9 $(24.4\times)$ & 5.0  & 3.99 $(1.1\times)$ \\
& 1D MC opt.     & 12,000 $(41.3\times)$ & 4.0 $(25.0\times)$  & 3.0  & -- \\
\midrule

\multirow{4}{*}{Dynamics}
& Bayesian GLM   & \textbf{280}     & \textbf{8.45}   & \textbf{0.19} & 1.6 \\
& GP-BO (TS)   & 4,400 $(15.7\times)$ & 48.5 $(5.7\times)$ & 1.51 & \textbf{1.2} $(0.7\times)$ \\
& GP-BO (qNEI)    & 5,100 $(18.2\times)$ & 49.0 $(5.8\times)$ & 1.0  & 1.49 $(0.9\times)$ \\
& 1D MC opt.     & 13,000 $(46.4\times)$ & 311.5 $(36.8\times)$ & 1.0  & -- \\
\midrule

\multirow{4}{*}{Non power-law}
& Bayesian GLM   & \textbf{280}     & --              & 2.3  & \textbf{0.56} \\
& GP-BO (TS)   & 3,500 $(12.5\times)$ & --           & \textbf{0.08} & 0.69 $(1.2\times)$ \\
& GP-BO (qNEI)    & 3,300 $(11.8\times)$ & -- & 2.0  & 0.70 $(1.2\times)$ \\
& 1D MC opt.     & 14,000 $(50.0\times)$ & --          & 1.0  & -- \\
\bottomrule
\end{tabular}
\end{table}

For the two GP-based BO baselines, we estimate the observation variance at each $\beta$ using 10 Monte Carlo replicates per evaluation and pass the resulting sample variance to the GP likelihood; using a single replicate leads to unstable noise estimates and degraded GP performance in this setting.
For numerical stability and to accommodate the strictly positive statistic and its large dynamic range, we perform GP modeling in log space and report results on the original scale.
Other parameters, such as initial points and batch size, are the same as in the proposed method.

\begin{figure}[!t]
\centering
\includegraphics[width=0.75\textwidth]{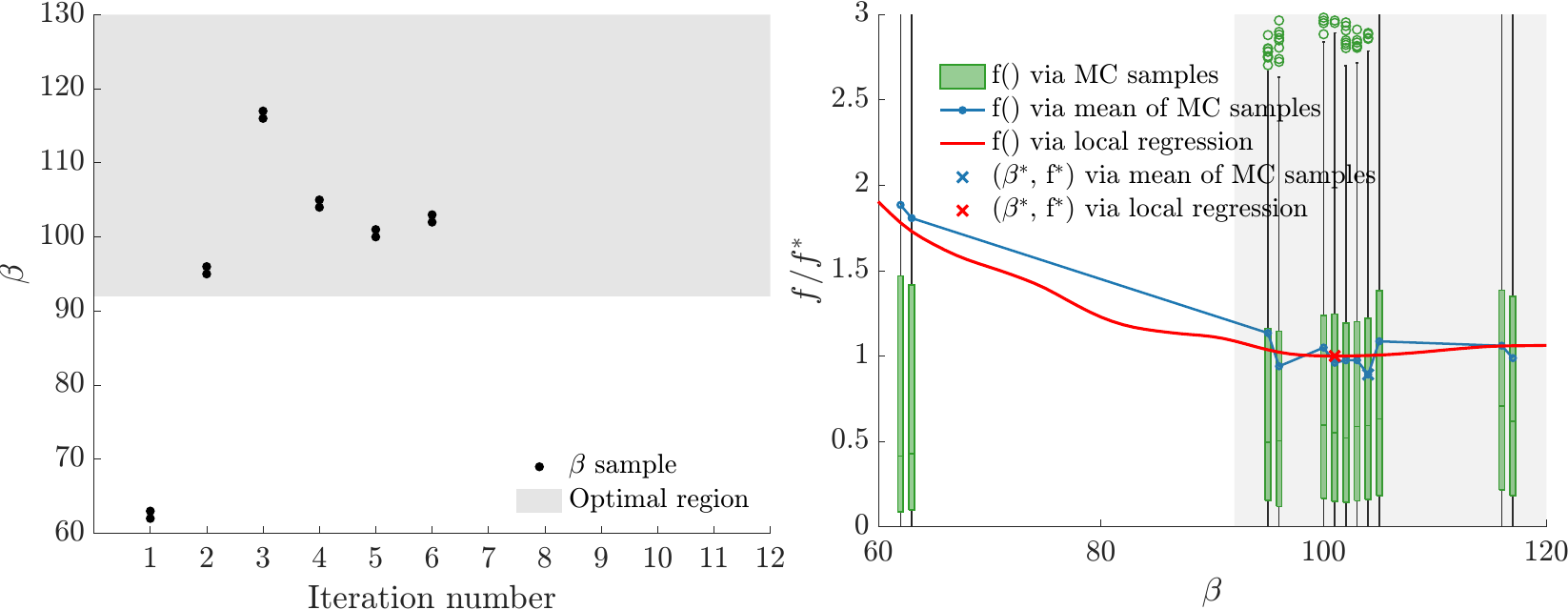}
\caption{Static problem: progression of one-dimensional Monte-Carlo based optimization. (Left) $\beta$ sample points during one-dimensional optimization iterations with 1000 MC samples.
(Right) Objective function versus $\beta$, with mean trends and local regression estimates.}
\label{fig:series-1D-ex1}
\end{figure}

\begin{figure}[!t]
\centering
\includegraphics[width=0.75\textwidth]{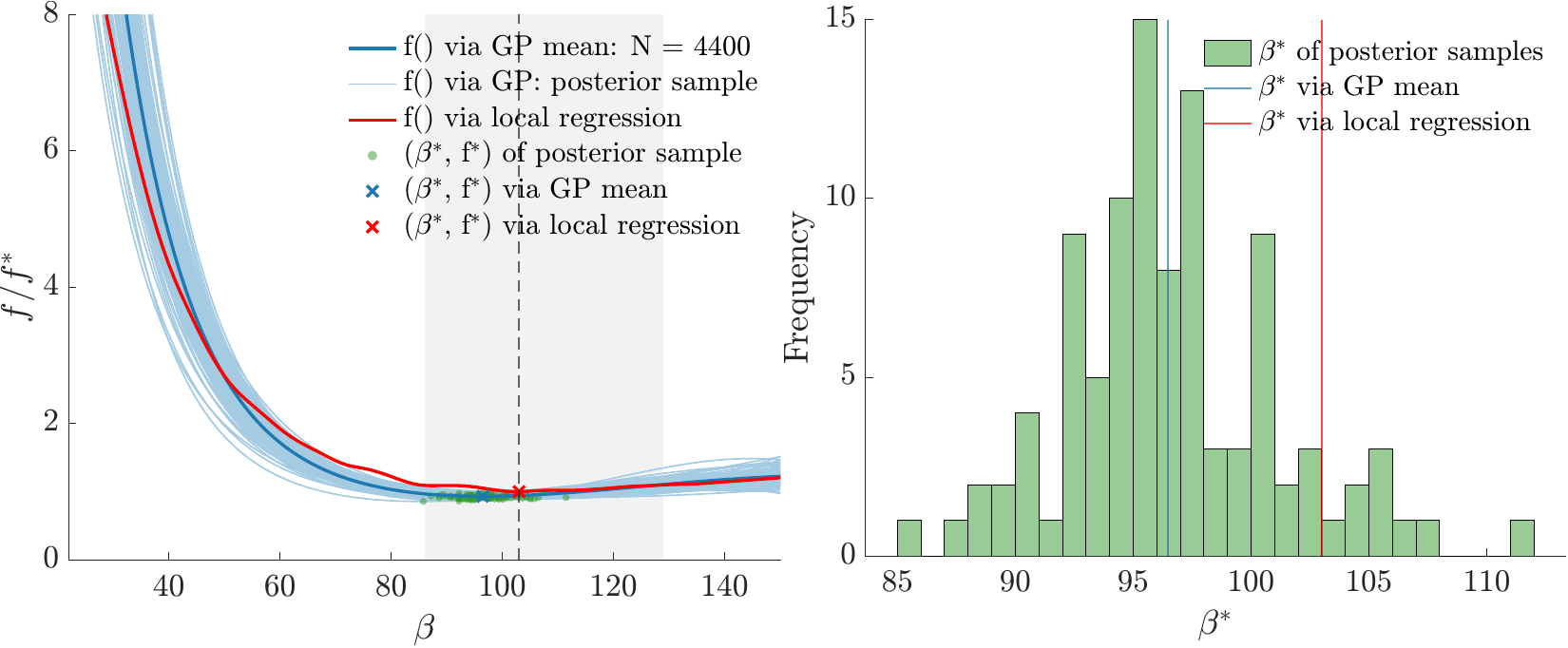}
\caption{Static problem: optimization results.
 (Left) $\beta^*$ obtained via GP surrogate with heteroscedastic noise and Thompson sampling acquisition function.
(Right) Histogram of $\beta^*$ posterior samples.}
\label{fig:f-gp-EX1}
\end{figure}

\begin{figure}[!t]
\centering
\includegraphics[width=0.65\textwidth]{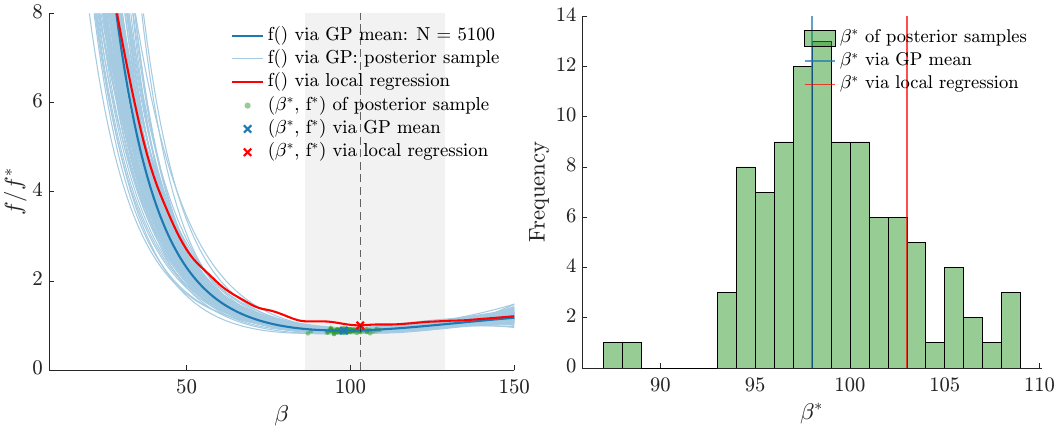}
\caption{Static problem: optimization results.
 (Left) $\beta^*$ obtained via GP surrogate with qNEI acquisition function.
(Right) Histogram of $\beta^*$ posterior samples.}
\label{fig:f-gp-qnei-EX1}
\end{figure}

\Cref{fig:f-glm-TS} shows some posterior samples of $f(\beta)$ at an iteration during the BO process,
along with the corresponding Thompson samples $(\beta^*, f^*)$, where $f^* := f(\beta^*)$. 
We observe that optimal $\beta$ values obtained via GLM, Bayesian GLM (see \cref{fig:f-glm-TS}), one-dimensional optimization scheme (see \cref{fig:series-1D-ex1}), and GP-based baselines (see \cref{fig:f-gp-EX1,fig:f-gp-qnei-EX1}) all fall into the optimal region.
This region is centered around the $\beta$ estimated via local regression and is defined as the range in which the mean square error does not vary by more than 10\%.
Mathematically, the optimal region is the set of $\beta$ values for which the locally smooth estimate of the mean objective $f(\beta)$ is within 10\% of its minimum, i.e., \{$\beta: f(\beta) \leq 1.1 f(\beta^*_{LR})$\}, where $\beta^*_{LR} = \argmin f(\beta)$ obtained via local regression.
Quantitatively, the optimal hyperparameter $\beta$ is 101 using the Bayesian GLM approach,
103 using local regression, 95 using GP surrogate with Thompson sampling acquisition function, 98 using GP surrogate with qNEI acquisition function, and 104 using one-dimensional optimization.
Although these estimates differ slightly, the impact on error characterization is minimal because all lie within the optimal region.
Beyond point estimates, the methods differ in the uncertainty of the identified optimum: the Bayesian GLM yields a tightly concentrated posterior distribution of $\beta^*$, whereas the GP-based BO baselines produce a much broader distribution of $\beta^*$ samples (see \cref{fig:f-glm-TS,fig:f-gp-EX1,fig:f-gp-qnei-EX1}).

Overall, the static example demonstrates that the proposed approach matches the accuracy of conventional one-dimensional Monte Carlo optimization and GP-based BO baseline while requiring $41\times$ fewer samples and achieving a $25\times$ speedup; it also uses $15\times$ fewer evaluations than the GP-based BO baselines in this setting.
We next consider a dynamic space-structure example to assess scalability in a much more computationally demanding regime.

\subsection{Dynamics problem: space structure}

\begin{figure}[!t]
    \centering
    \includegraphics[width=0.75\textwidth]{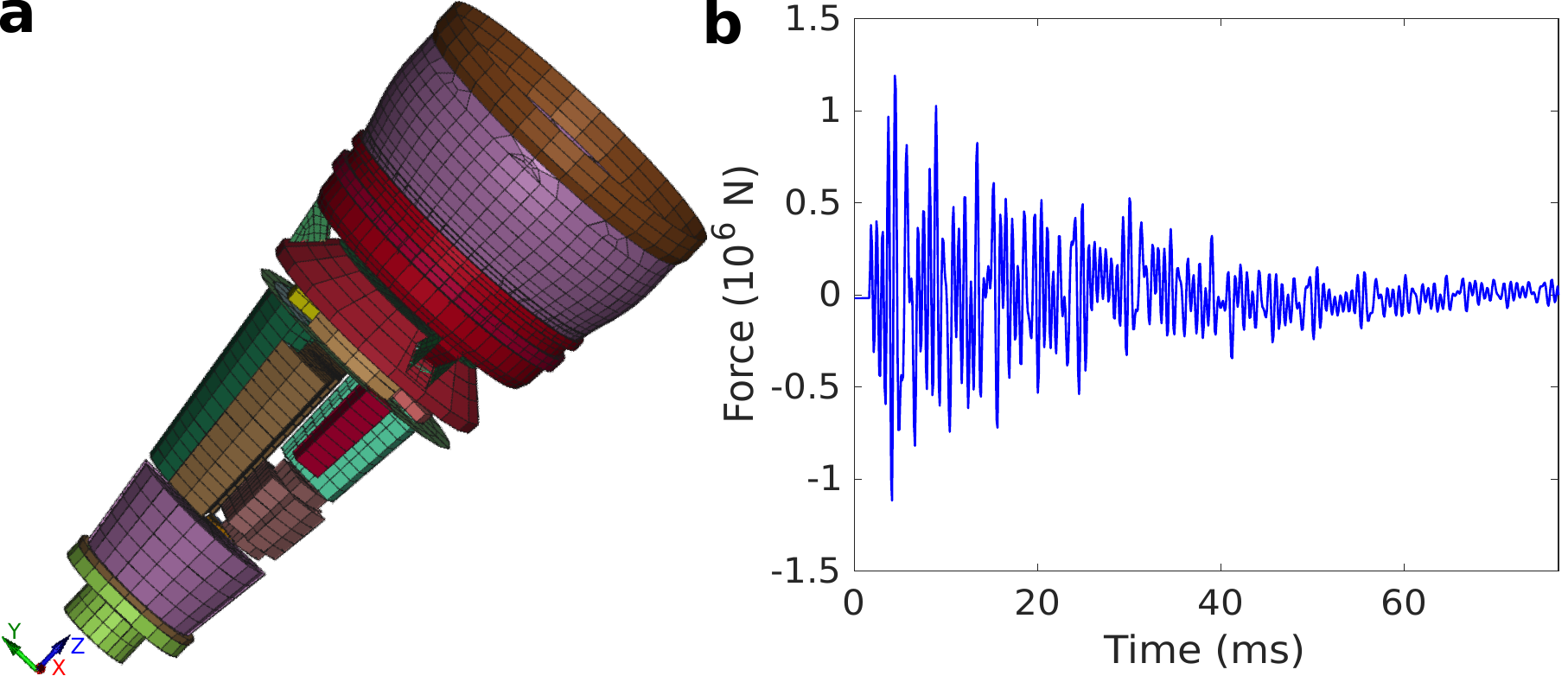}
    \caption{Dynamics problem: (a) space structure; (b) loading.}
    \label{fig:space-structure}
\end{figure}

We consider a dynamic problem involving a component of a space structure subjected to impulse loading in the $z$-direction, as illustrated in \cref{fig:space-structure}.
The structure consists of two parts: an upper section with a heavy central mass (approximately 100 times heavier than the rest of the structure), connected to the sidewalls via rigid links, and a lower section that houses essential components, such as a shock-absorption block.
As shown in \cref{fig:space-structure}b, the impulse load is applied at the center of mass in the $z$-direction.  
The load propagates to the upper section through the sidewalls and to the lower section via the mounting pedestal.
The functionality of critical components in the lower section may be compromised under impulse loading, necessitating monitoring of their dynamic response.
The quantity of interest (QoI) is taken as the $x$-velocity at one of the critical nodes of the essential components. 
The space structure is modeled as free-floating, with no external boundary conditions.
The dynamics of the structure are governed by a second-order ordinary differential equation (ODE) system with 42,486 DoFs:
\begin{equation}\label{eq:HDM-space}
    \mathbf{M} \ddot{\mathbf{x}}(t) + \mathbf{C} \dot{\mathbf{x}}(t) + \mathbf{K} \mathbf{x}(t) = \mathbf{f}(t) ,\quad t \in [t_0,T]
\end{equation}
with the initial conditions $\mathbf{x}(t_0) = 0$ and $\dot{\mathbf{x}}(t_0) = 0$.
The damping matrix $\mathbf{C}$ is proportional to the stiffness, with a proportionality coefficient of $6.366 \times 10^{-6}$. 
The HDM is solved using the Newmark-$\beta$ time integration scheme
with a time step of $5 \times 10^{-2}$ ms.
Each HDM simulation requires approximately 38 minutes of computational time, making repeated evaluations infeasible for tasks such as optimization or uncertainty quantification.
To reduce cost, we construct a ROM using proper orthogonal decomposition (POD) with a reduced dimension $k = 10$.
For $k = 10$, the retained POD modes yield a relative truncation error of $10^{-8}$.
The governing equation of the ROM is given by:
\begin{equation}
  \mathbf{M}_k \ddot{\mathbf{q}}(t) + \mathbf{C}_k \dot{\mathbf{q}}(t) + \mathbf{K}_k \mathbf{q}(t) = \mathbf{V}^\intercal \mathbf{f}(t),
  \quad \mathbf{x}_k(t) = \mathbf{V}\mathbf{q}(t) , \quad t \in [t_0,T]
\end{equation}
where, $\mathbf{M}_k = [\mathbf{V}]^\intercal [\mathbf{M}] [\mathbf{V}], \mathbf{C}_k = [\mathbf{V}]^\intercal \mathbf{C} [\mathbf{V}],$ and $\mathbf{K}_k = [\mathbf{V}]^\intercal \mathbf{K} [\mathbf{V}]$. 
The ROM requires only 0.2 seconds, making it 11,200 times faster than the HDM.
However, it introduces mode-truncation error, which must be characterized before practical use. 
We address the mode-truncation error with the SROM framework.
The SROM retains the same governing form as the ROM but replaces the deterministic basis $\mathbf{V}$ with a stochastic basis $\mathbf{W}$ sampled from the stochastic subspace model of \citet{yadav2025ss}.

The hyperparameter $\beta$ associated with the stochastic subspace model is estimated
by minimizing the objective function defined in \cref{eq:objective}. In this example, $d_o(\mathbf{u}_E)$ is the $L^2$ distance between HDM and ROM velocity at a critical node, 
and $d_o(\mathbf{u}_L)$ is the $L^2$ distance between ROM and SROM velocity at the same node.
We estimate the hyperparameter $\beta$ using the methodology discussed in \cref{sec:methodology,sec:SROM} and compare the results with those from a one-dimensional optimization algorithm with Monte Carlo samples and GP-based baselines.

\begin{figure}[!tb]
    \centering
    \includegraphics[width=0.75\textwidth]{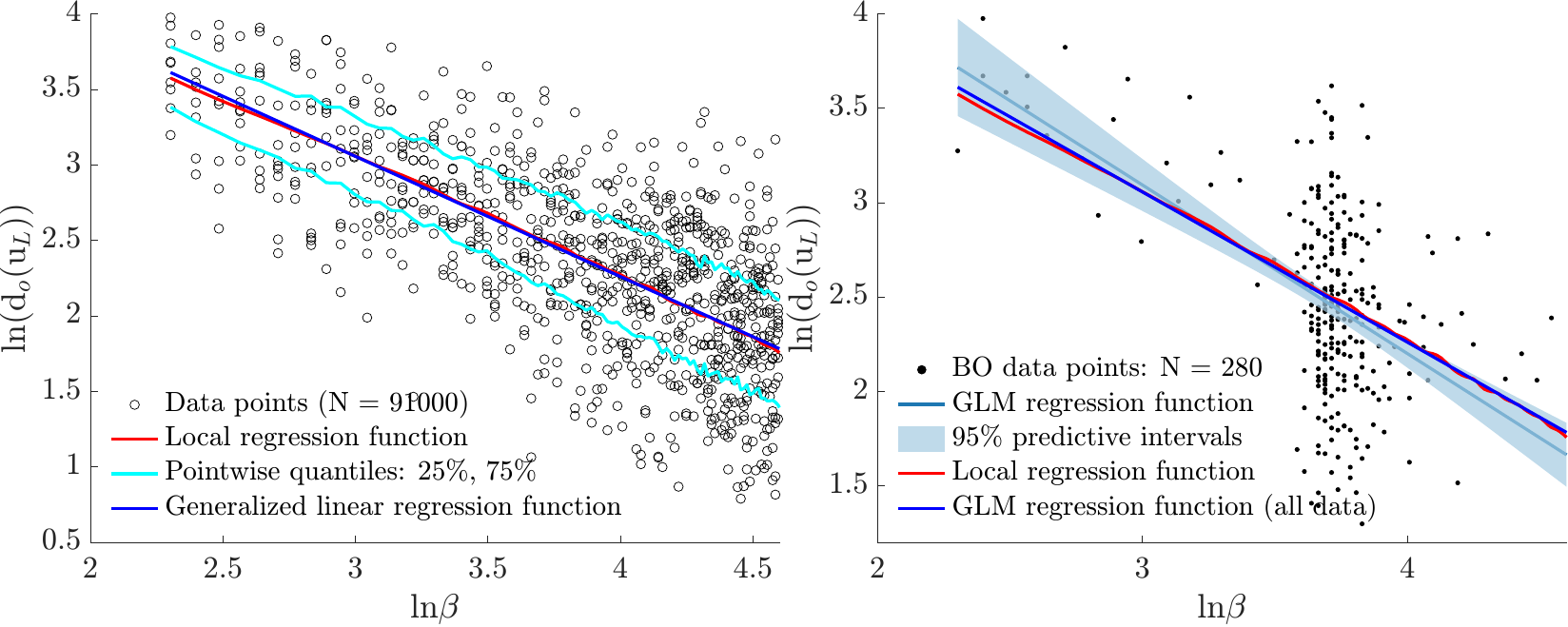}
    \caption{Dynamics problem: raw data and surrogate model fit.
    (Left) Raw data is highly noisy with a downward linear trend.
    Plotted points are 1\% of the data, uniformly subsampled.
    (Right) Bayesian GLM captures the GLM regression function accurately with quantified error,
    using much fewer data points.}
    \label{fig:glm-Bayes-EX2}
\end{figure}

\begin{figure}[!tb]
\centering
\includegraphics[width=0.75\textwidth]{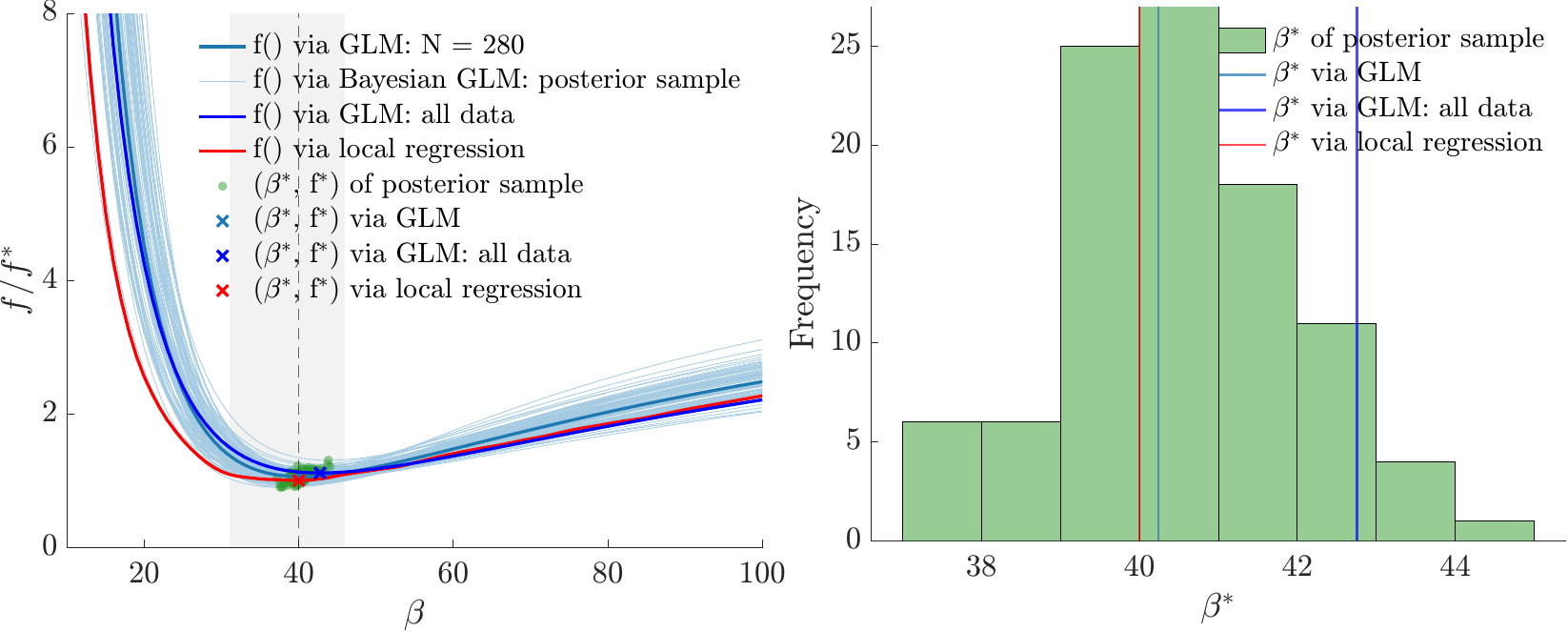}
\caption{Dynamics problem: optimization results under different approaches. (Left) Posterior distribution of $(\beta^*, f^*)$ well captures the true statistics.
(Right) Histogram of $\beta^*$ posterior samples.}
\label{fig:f-glm-TS-EX2}
\end{figure}

\begin{figure}[!tb]
\centering
\includegraphics[width=0.75\textwidth]{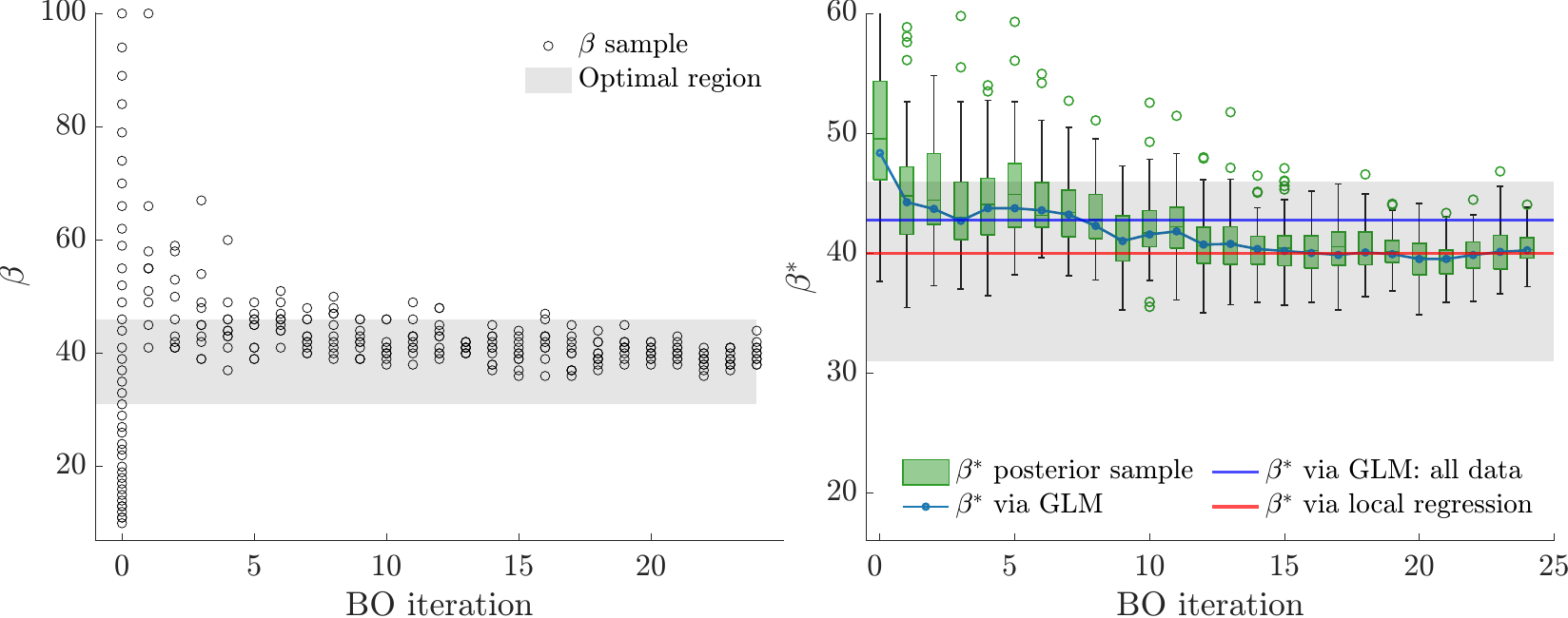}
\caption{Dynamics problem: progression of Bayesian optimization. (Left) Sample points during BO iterations.
(Right) Evolution of $\beta^*$ posterior distribution and GLM estimate.}
\label{fig:series-BO-EX2}
\end{figure}

\begin{figure}[!tb]
\centering
\includegraphics[width=0.75\textwidth]{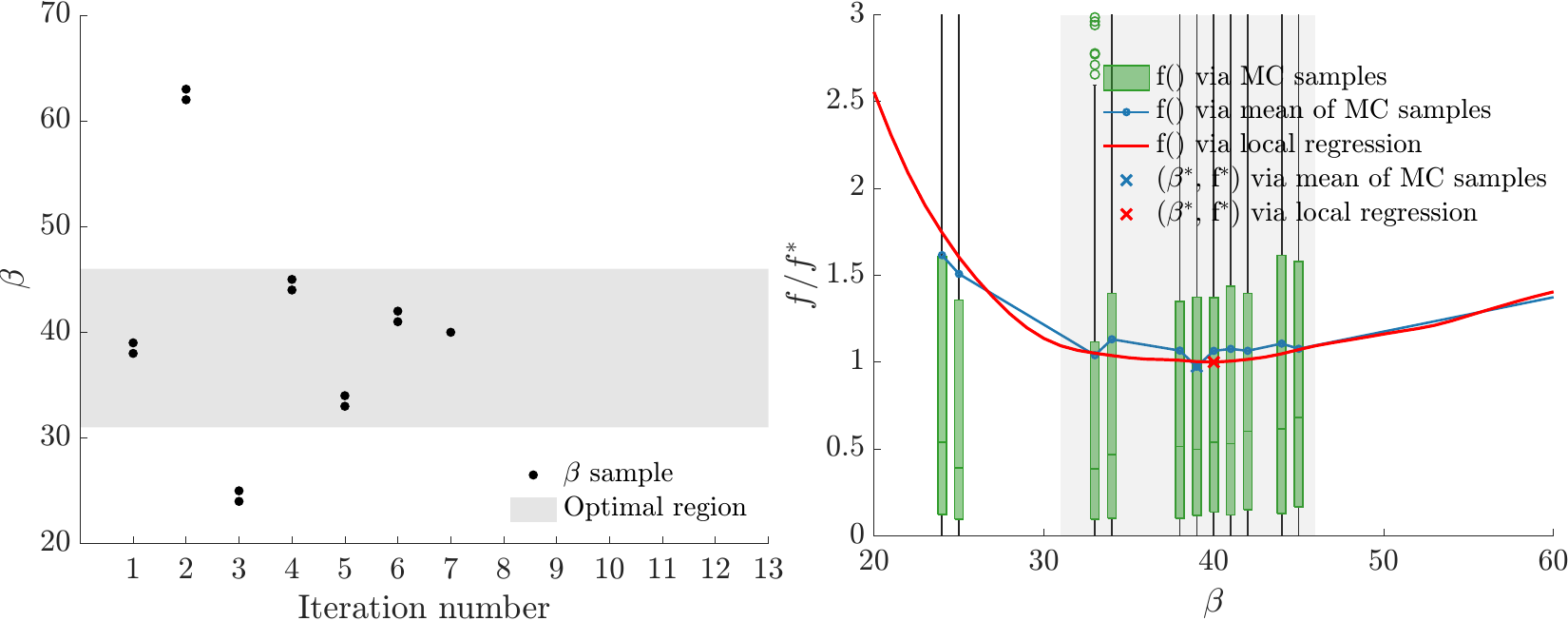}
\caption{Dynamics problem: progression of one-dimensional Monte-Carlo based optimization. (Left) $\beta$ sample points during one-dimensional optimization iterations with 1000 MC samples.
(Right) Objective function versus $\beta$, with mean trends and local regression estimates.}
\label{fig:series-1D-ex2}
\end{figure}

\begin{figure}[!tb]
\centering
\includegraphics[width=0.75\textwidth]{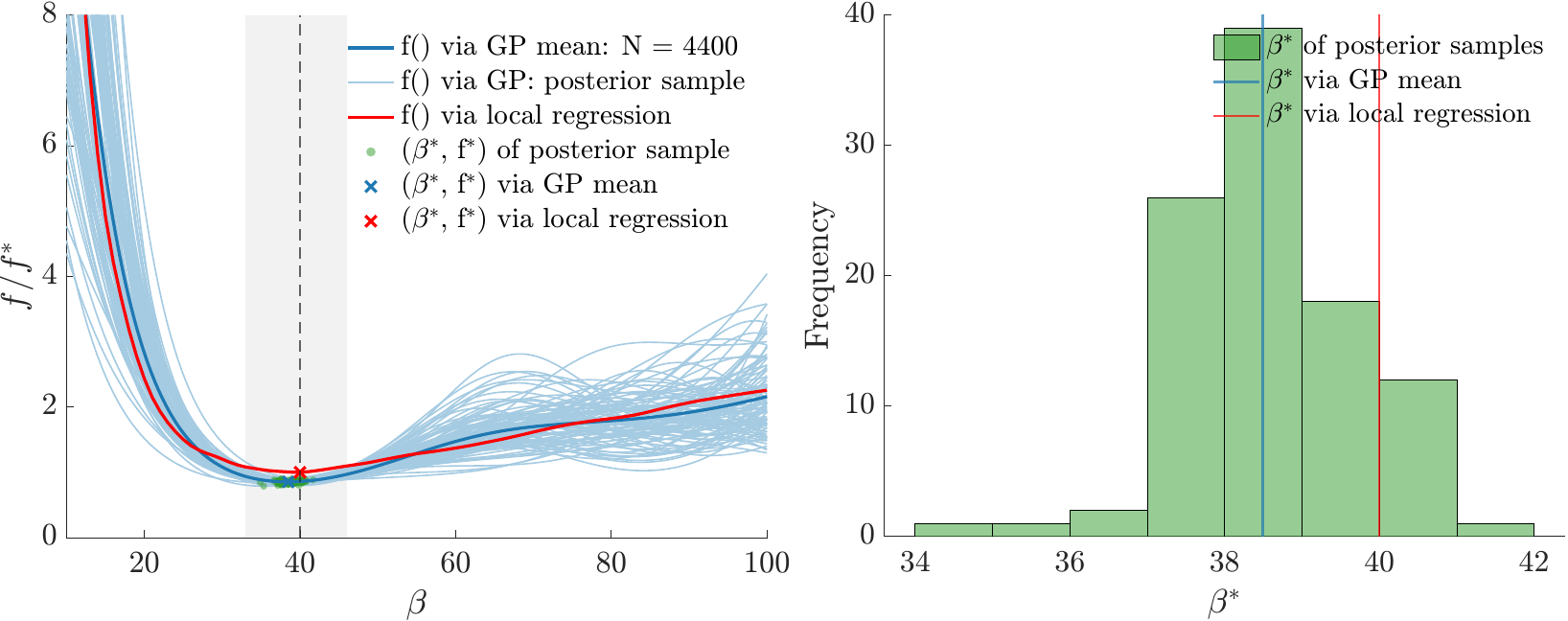}
\caption{Dynamic problem: optimization results.
 (Left) $\beta^*$ obtained via GP surrogate with heteroscedastic noise and Thompson sampling acquisition function.
(Right) Histogram of $\beta^*$ posterior samples.}
\label{fig:f-gp-EX2}
\end{figure}

\begin{figure}[!tb]
\centering
\includegraphics[width=0.65\textwidth]{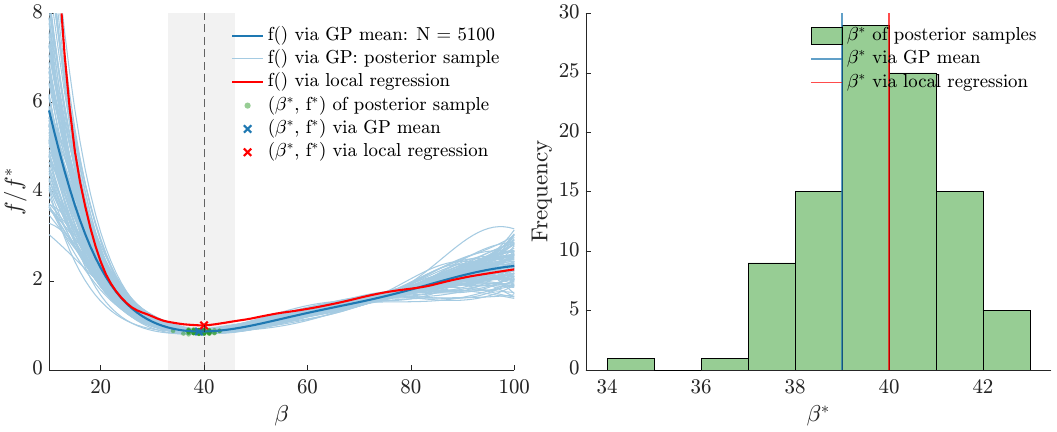}
\caption{Dynamic problem: optimization results.
 (Left) $\beta^*$ obtained via GP surrogate with qNEI acquisition function.
(Right) Histogram of $\beta^*$ posterior samples.}
\label{fig:f-gp-qnei-EX2}
\end{figure}

\Cref{fig:glm-Bayes-EX2} shows that the Bayesian GLM surrogate accurately captures the expected power-law scaling even from highly noisy raw data, while \cref{fig:f-glm-TS-EX2} and \cref{fig:series-BO-EX2} indicate rapid convergence of the posterior distribution of $\beta^*$.
The Bayesian GLM requires only 280 evaluations to estimate the optimal concentration parameter, compared with 13,000 evaluations for one-dimensional Monte Carlo–based optimization (see \cref{tab:all-examples} and \cref{fig:series-1D-ex2}), 4,400 evaluations for the GP-based BO with Thompson acquisition function (see \cref{fig:f-gp-EX2}), and 5,100 evaluations for GP-based BO with qNEI acquisition function (see \cref{fig:f-gp-qnei-EX2}).
In terms of computational resources, this corresponds to a 46-fold reduction in data usage and a 36-fold reduction in runtime compared to a one-dimensional optimization scheme, and around 15-fold reduction in data usage compared to the GP-based BO baselines.
Across all methods, the identified optima lie within the same optimal region.
Quantitatively, the estimated optima are $\beta=40$ (Bayesian GLM), $\beta=40$ (local regression), $\beta=39$ (GP-based baselines), and $\beta=39$ (one-dimensional Monte Carlo optimization), with negligible impact on error characterization.
For the GP-based BO baseline, the posterior sample standard deviation ($\sigma(\beta^*)$) can fluctuate across iterations and may not decrease monotonically.
To obtain a stable result, we adopt a stopping criterion: the algorithm stops when $\sigma(\beta^*)$ remains below a prescribed threshold for three consecutive iterations.

Across both examples, the Bayesian GLM framework delivers accurate estimates of the optimal concentration parameter with substantial savings in data and runtime.
Conventional one-dimensional optimization relies on Monte Carlo estimates at each candidate $\beta$, yielding noisy objective evaluations and high sample costs.
GP-based BO baselines also require a few Monte Carlo samples to estimate data noise.
By contrast, the Bayesian GLM surrogate (i) exploits the expected log--log linear (power-law) scaling with respect to $\beta$, (ii) enables analytic evaluation of the expectation in the objective, and (iii) concentrates evaluations in informative regions through Bayesian optimization.

\section{Discussion} \label{sec:discussion}

The methodology we propose in this paper follows analytically from the surrogate model in \cref{eq:glm},
and Bayesian optimization via Thompson sampling has strong theoretical guarantees on their convergence \citep{Russo2014}.
The performance of our approach, therefore, rests on the validity and accuracy of the surrogate model.
Here we discuss these issues in detail.

\subsection{Conditional residual model}

\begin{figure}[!t]
\centering
\includegraphics[width=0.75\textwidth]{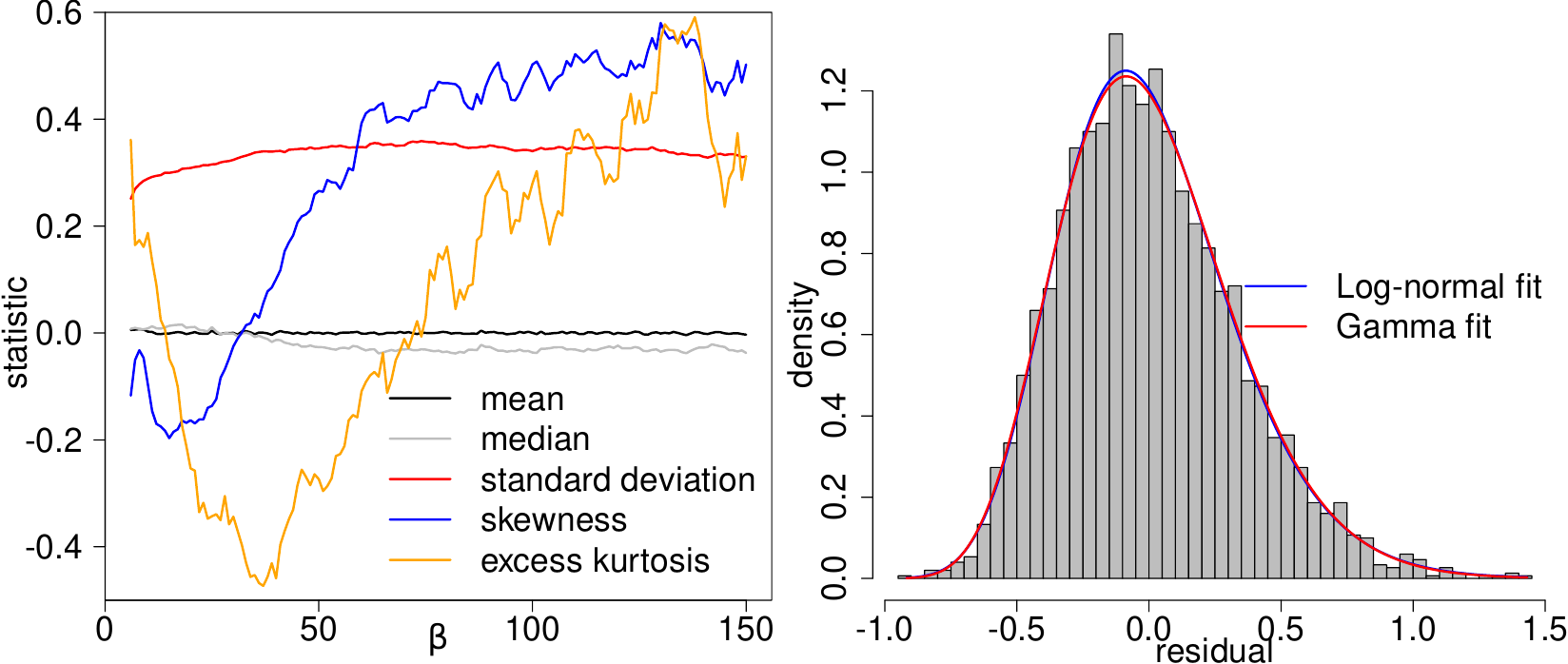}
\caption{Diagnosis of residual distributions.
(Left) Statistics of the residual distribution conditional on $\beta$.
(Right) Histogram of the residual distribution for $\beta = 150$, and the best-fit PDFs using log-normal and Gamma distributions.}
\label{fig:residual}
\end{figure}

The Bayesian GLM in \cref{eq:glm} can be broken apart into three aspects:
conditional mean function, conditional residual distribution, and prior on the model parameters.
We validated the conditional mean model in \cref{fig:glm-Bayes} as well as in \cref{fig:glm-Bayes-EX2}.
Now we examine the conditional residual model, which is a stationary Gaussian white noise.
To test stationarity, we compute summary statistics of the residual distributions:
mean, median, standard deviation, skewness, and excess kurtosis.
For each integer $\beta$ value, 1000 sample points are used to estimate the statistics.
The results for the static problem are shown in \cref{fig:residual}, smoothed with a rolling window of width 6.
The sample mean of the residual, as expected, is close to zero.
The median and standard deviation are largely stable for $\beta > 50$.
A stable variance means that homoscedasticity is a valid assumption.
However, skewness and kurtosis do not stabilize in the range of $\beta$ we inspect,
which means that stationarity does not hold.

To test the distributional assumption, we simply plot the histogram of the sample residual, see \cref{fig:residual}.
While we assume Gaussian distribution, the motivating example in \cref{eq:glm-example} points to $\ln |z|$,
where $z$ is a standard Gaussian random variable.
On the other hand, fitting the data to classic parametric families of distributions
suggests that shifted log-normal and Gamma distributions are good fits.
These two distributions, however, have skewness and kurtosis tied to the log-scale and the shape parameter, respectively,
making them not a robust choice across the $\beta$ values.

Using informative priors on the model parameters can further reduce the sample size needed for an accurate surrogate model,
but this requires further knowledge about the variable $s$ and $\beta$.
If $\beta$ behaves like sample size and $s$ a norm, then from \cref{eq:glm-example}
we may guess that the coefficient $a$ is around $\frac{1}{2}$ and the intersect $\ln b$ is around $\ln \sigma$,
and therefore impose an informative prior on $(a, \ln b, \varepsilon^2)$ as:
$a \sim N(-\frac{1}{2}, \sigma_a^2)$ and $\ln b \sim N(\ln \sigma, \sigma_b^2)$.
In fact, in the two examples demonstrated in the work,
the estimated values of $a$ are -0.5798 (\cref{fig:glm-Bayes}) and -0.7958 (\cref{fig:glm-Bayes-EX2}), respectively.
However, even for the classical example,
the signs and values of the parameters vary significantly based on the exact formulation of $s$ and $\beta$.
Imposing a strong prior thus risks steering the model in a wrong direction, undermining sample efficiency.
Given that the standard noninformative prior performs well and is easy to work with,
we do not pursue informative priors further in this work.

\subsection{Power-law assumption} \label{ex3}
The power-law assumption \cref{eq:assumption} applies when, for example,
the summary statistic is a vector $p$-norm, as is the case in our numerical examples.
To assess the robustness of our method when this assumption is violated,
we consider a synthetic example in which the relationship between $s(\omega)$ and $\beta$
is intentionally nonlinear on the log--log scale.
We define the objective function as:
\begin{equation} \label{eq:example3}
    \ftrue(\beta) = \mathbb{E}[(s(\omega) - s_o)^2 \mid \beta],
\end{equation}
where, $\log(s(\omega)) = -\log(\log(\beta)) + \epsilon z, z \sim N(0,1)$. 
This example mimics the structure of the previous two examples,
with a different scaling law between $s$ and $\beta$.
We set $s_o = 0.3$ and $\epsilon = 0.1$.
This construction preserves the overall optimization-under-uncertainty structure while violating the power-law (log--log linear) assumption underlying the GLM surrogate in \cref{eq:glm}.

\begin{figure}[!tb]
    \centering
    \includegraphics[width=0.75\textwidth]{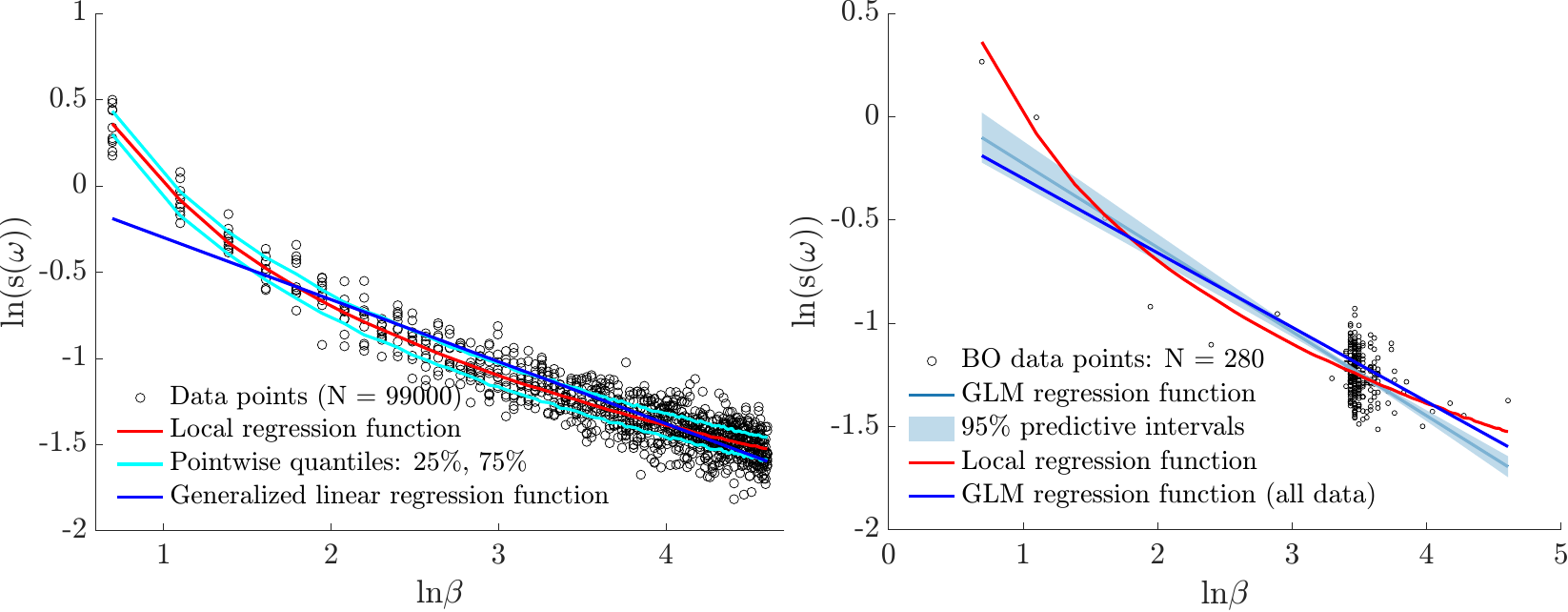}
    \caption{Non-power-law problem: raw data and surrogate model fit.
    (Left) Raw data is highly noisy with a downward linear trend.
    Plotted points are 1\% of the data, uniformly subsampled.
    (Right) Bayesian GLM captures the local regression function accurately
    near the optima using much fewer data points.}
    \label{fig:glm-Bayes-EX3}
\end{figure}

\begin{figure}[!tb]
\centering
\includegraphics[width=0.65\textwidth]{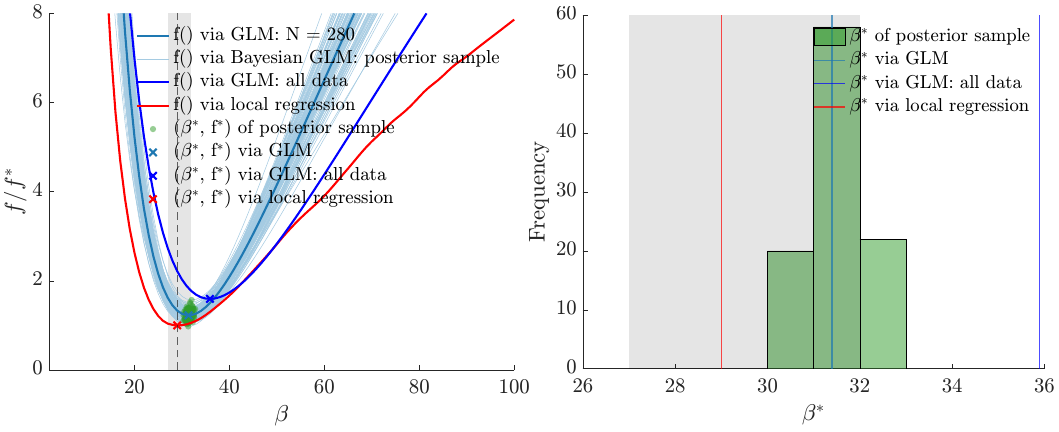}
\caption{Non-power-law problem: optimization results under different approaches. (Left) Posterior distribution of $(\beta^*, f^*)$ well captures the true statistics.
(Right) Histogram of posterior samples of $\beta^*$, indicating residual bias due to posterior mass falling outside the optimal region.}
\label{fig:f-glm-TS-EX3}
\end{figure}

\Cref{fig:glm-Bayes-EX3} confirms that the global trend is not well represented by a single log--log linear model, indicating surrogate misspecification.
Such misspecification usually creates bias.
However, given the monotonicity of the underlying objective function, we can argue that, even though the GLM surrogate has bias, the deviation is bounded to some extent.
That is, the GLM bias is constrained by the monotonicity of the objective function.
The Bayesian GLM can provide a useful local approximation near the optimum, which is sufficient for effective acquisition decisions.
As shown in \cref{fig:f-glm-TS-EX3}, the Thompson-sampled surrogates yield $\beta^*$ samples concentrated near the optimal region identified by local regression.

\begin{figure}[!tb]
\centering
\includegraphics[width=0.75\textwidth]{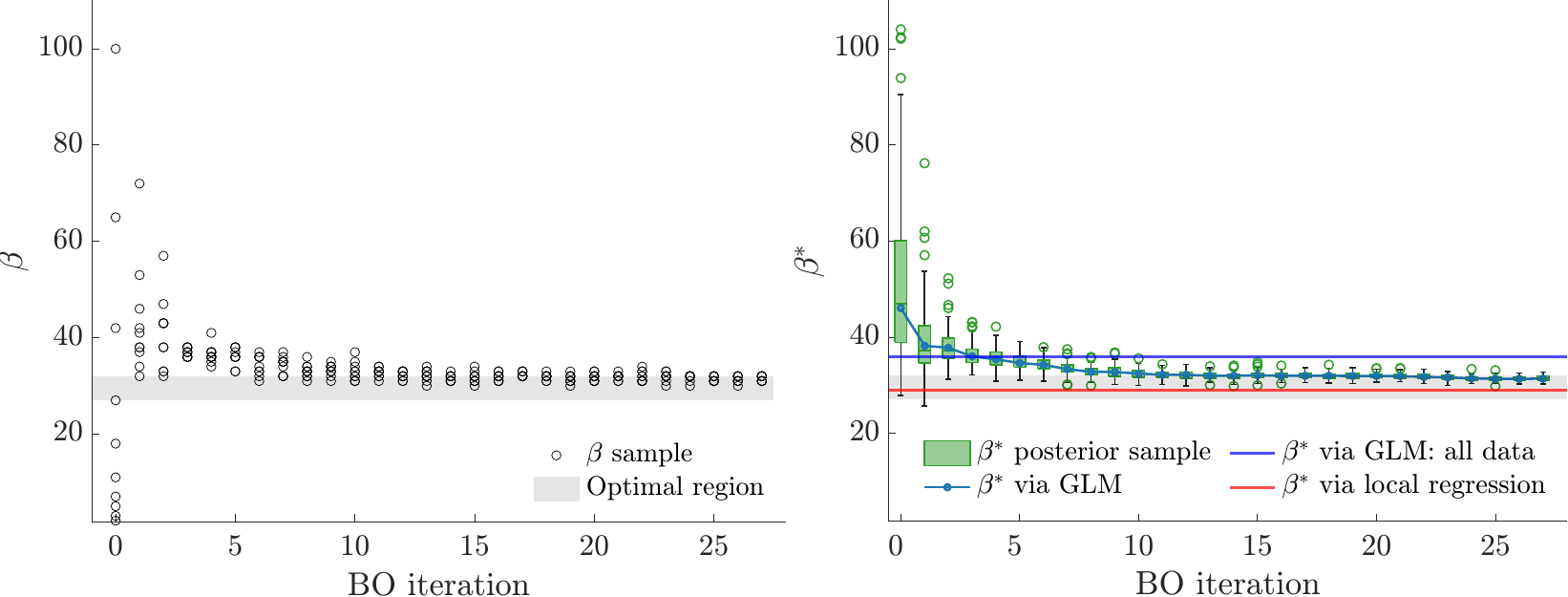}
\caption{Non-power-law problem: progression of Bayesian optimization. (Left) Sample points during BO iterations.
(Right) Evolution of $\beta^*$ posterior distribution and GLM estimate.}
\label{fig:series-BO-EX3}
\end{figure}

\begin{figure}[!tb]
\centering
\includegraphics[width=0.75\textwidth]{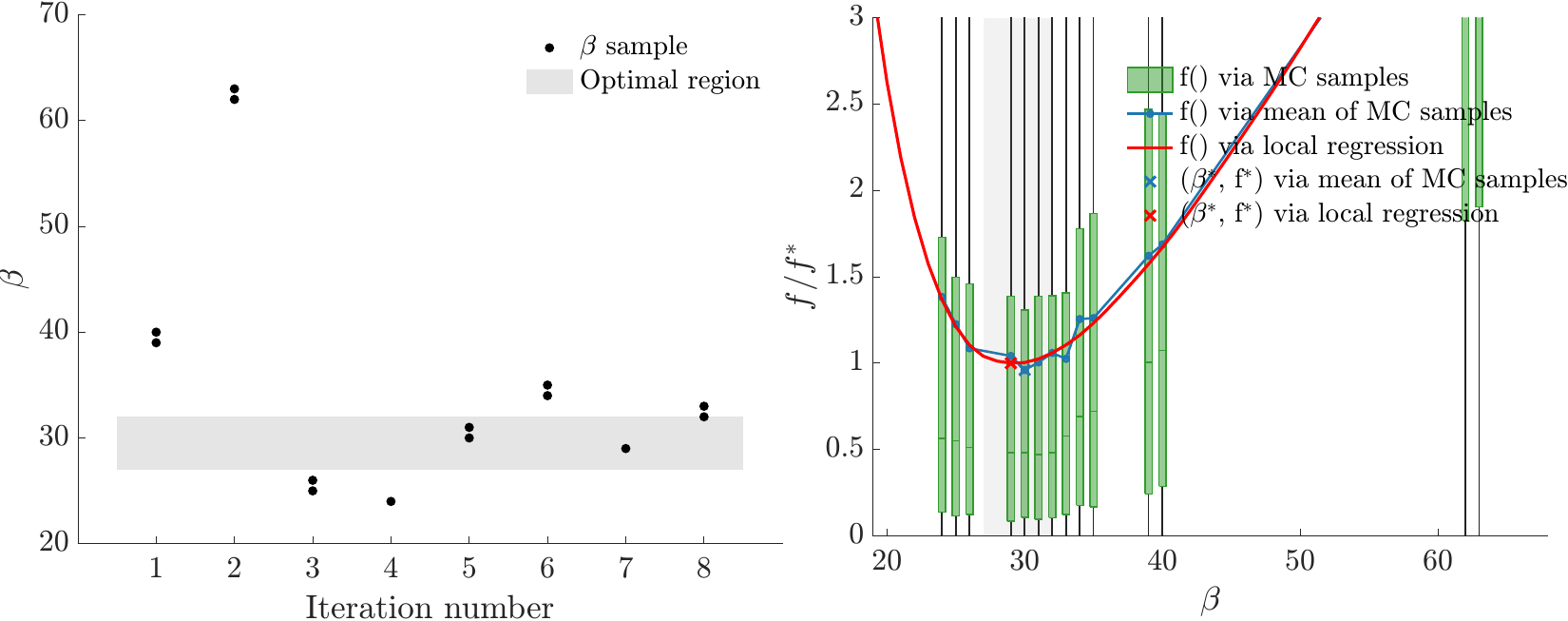}
\caption{Non-power-law problem: progression of one-dimensional Monte-Carlo based optimization. (Left) $\beta$ sample points during one-dimensional optimization iterations with 1000 MC samples.
(Right) Objective function versus $\beta$, with mean trends and local regression estimates.}
\label{fig:series-1D-ex3}
\end{figure}

\begin{figure}[!tb]
\centering
\includegraphics[width=0.75\textwidth]{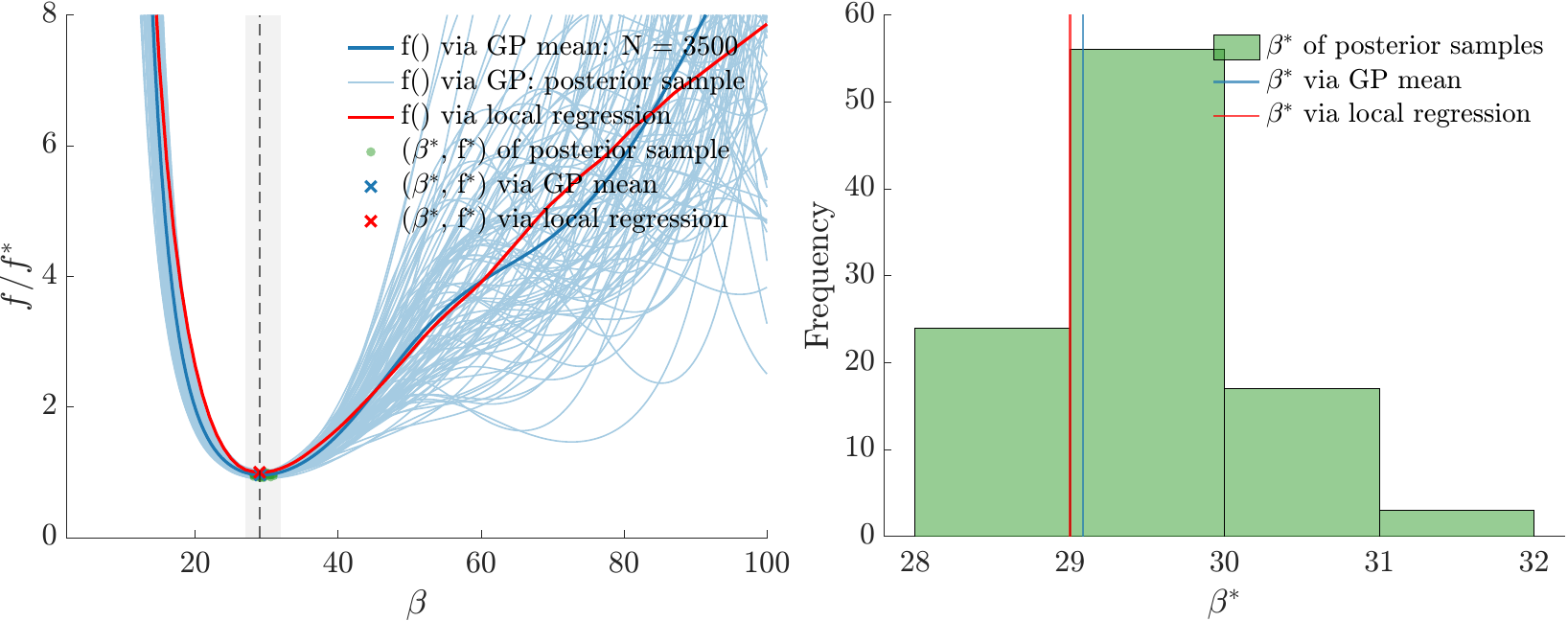}
\caption{Non-power-law problem: optimization results.
 (Left) $\beta^*$ obtained via GP surrogate with heteroscedastic noise and Thompson sampling acquisition function.
(Right) Histogram of $\beta^*$ posterior samples.}
\label{fig:f-gp-EX3}
\end{figure}

\begin{figure}[!tb]
\centering
\includegraphics[width=0.65\textwidth]{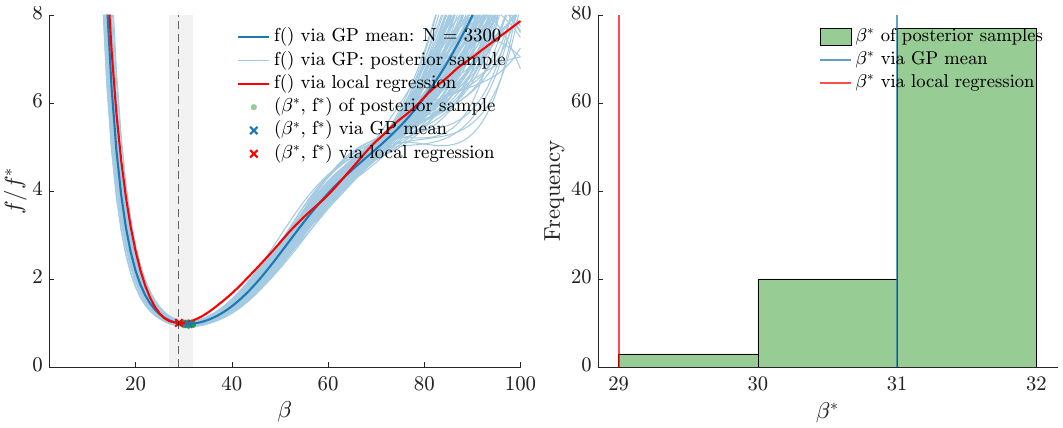}
\caption{Non-power-law problem: optimization results.
 (Left) $\beta^*$ obtained via GP surrogate with qNEI acquisition function.
(Right) Histogram of $\beta^*$ posterior samples.}
\label{fig:f-gp-qnei-EX3}
\end{figure}

\cref{fig:series-BO-EX3} and \cref{fig:series-1D-ex3} further illustrate the difference in sampling behavior: the proposed method concentrates evaluations near the optimum, whereas one-dimensional optimization requires repeated Monte Carlo estimation at each candidate $\beta$.
Across methods, the estimated optima remain consistent.
Specifically, the optimal hyperparameter is $\beta=31$ using the Bayesian GLM approach, $\beta=30$ using one-dimensional Monte Carlo optimization (see \cref{fig:series-1D-ex3}), $\beta=29$ using local regression and GP-based BO with Thompson sampling acquisition function (see \cref{fig:f-gp-EX3}), and 31 using GP-based BO with qNEI acquisition function (see \cref{fig:f-gp-qnei-EX3}.
As in other examples, we observe that Bayesian GLM achieves optimal results, with substantial gains in data efficiency, even though the power-law assumption is violated  
Thus, the example demonstrates that even though the Bayesian GLM method suffers a loss of surrogate fidelity under model misspecification, this loss is not catastrophic for the accuracy of the selected hyperparameter.

\subsection{Ablation study}\label{sec:ablation-study}
\paragraph{Sensitivity to initialization:} We examine the sensitivity of the proposed method to the choice of the initial dataset.
Because the Bayesian GLM surrogate is first fit using the initial observations, different initial $\beta$ locations may affect early estimates of the scaling parameters and, consequently, the subsequent BO trajectory.

For each example, we keep the optimization domain $[\beta_{\min},\beta_{\max}]$, the initial design size $n_0$, the batch size $B$, and all other algorithmic settings fixed.
We vary only the initial sampling design by shifting the lower bound used to generate the \emph{initial} $\beta$ values to $\beta_{\min}\in\{k,\;k+5,\;k+10\}$, while keeping the optimization domain unchanged. That is, only the initial design is modified; all subsequent BO iterations optimize over the same domain $[\beta_{\min},\beta_{\max}]$.
This procedure produces three distinct initial datasets without changing the underlying problem.

For each run, we record the final estimate $\hat{\beta}$ and the total number of objective function evaluations required to meet the stopping criterion.
\Cref{tab:ablation} summarizes the results.
Across all three examples, the estimated optima remain stable (that is, all fall within the optimal region) across runs, and the evaluation counts do not vary much, indicating that the method is robust to the choice of initialization.

\begin{table}[!t]
\centering
\caption{Initialization sensitivity study across three examples. Three runs correspond to different initial datasets (all other settings fixed).}
\label{tab:ablation}
\begin{tabular}{llrrr}
\toprule
Example & Metric & Run 1 & Run 2 & Run 3 \\
\midrule
\multirow{2}{*}{Static}
& $\hat{\beta}$ & 98 & 98 & 97 \\
& \# evaluations & 280 & 290 & 290 \\
\midrule
\multirow{2}{*}{Dynamics}
& $\hat{\beta}$ & 40 & 43 & 43 \\
& \# evaluations & 280 & 290 & 290 \\
\midrule
\multirow{2}{*}{Non power-law}
& $\hat{\beta}$ & 31 & 30 & 30 \\
& \# evaluations & 300 & 300 & 290 \\
\bottomrule
\end{tabular}
\end{table}

\paragraph{Multi-seed statistical evaluation.}
Because both the stochastic objective evaluations and the optimization procedures considered in this work involve randomness, conclusions drawn from a single run are not sufficient to assess robustness.
We therefore repeat each method over $N_{\mathrm{seed}} = 15$ independent random seeds and report summary statistics across runs.
For each method, we use the same search domain, initialization rule, stopping criterion, and total evaluation budget.
The random seed affects the initial design, posterior samples, Monte Carlo draws, and any randomized components of the acquisition-function optimization.
In this way, observed performance differences reflect the behavior of the methods rather than a favorable random realization.
For each run, we record the final estimated optimal hyperparameter, $\hat{\beta}^*$, the posterior spread $\sigma(\beta^*)$, and the number of objective function evaluations required to satisfy the prescribed stopping criterion.
We then report the mean and standard deviation of these quantities across seeds.

\Cref{tab:multiseed} summarizes the multi-seed results.
Overall, the proposed Bayesian GLM approach yields stable estimates of $\hat{\beta}^*$ across repeated runs while requiring substantially fewer data points than the GP-based BO baselines.
These results indicate that the method is robust to random initialization and remains statistically competitive across the three examples.
The multi-seed analysis shows that the observed performance is not due to a single favorable run.
Instead, the proposed method consistently achieves accurate, data-efficient optimization across repeated experiments, which is especially important in the present optimization-under-uncertainty setting, where objective function noise and randomized optimization components can otherwise obscure method comparisons.

\begin{table}[!t]
\centering
\caption{Multi-seed statistical evaluation over $N_{\mathrm{seed}}=15$ independent runs.
For each method and example, we report the mean and standard deviation of the estimated optimal hyperparameter, the standard deviation of the posterior samples, $\sigma(\beta^*)$, and the mean number of data points.}
\label{tab:multiseed}
\small
\begin{tabular}{llccc}
\toprule
Example & Method & Mean $\hat{\beta}^* \pm$ std & Mean $\sigma(\beta^*) \pm$ std & Mean Data Points \\
\midrule

\multirow{3}{*}{Static ($\beta^*_{LR}=103$)}
& Bayesian GLM   & $97.9 \pm 1.63$  & $3.82 \pm 0.27$ & $309$  \\
& GP-BO (TS)     & $101.3 \pm 6.26$ & $6.44 \pm 3.39$ & $5100$ \\
& GP-BO (qNEI)   & $97.2 \pm 5.37$  & $6.94 \pm 4.04$ & $5100$ \\
\midrule

\multirow{3}{*}{Dynamics ($\beta^*_{LR}=40$)}
& Bayesian GLM   & $41.8 \pm 1.34$  & $1.78 \pm 0.20$ & $340$  \\
& GP-BO (TS)     & $37.2 \pm 1.66$  & $1.32 \pm 0.46$ & $4727$ \\
& GP-BO (qNEI)   & $38.5 \pm 2.80$  & $1.45 \pm 0.68$ & $4407$ \\
\midrule

\multirow{3}{*}{Non power-law ($\beta^*_{LR}=29$)}
& Bayesian GLM   & $49.1 \pm 1.35$  & $3.03 \pm 0.26$ & $273$  \\
& GP-BO (TS)     & $44.5 \pm 6.30$  & $7.28 \pm 2.70$ & $5100$ \\
& GP-BO (qNEI)   & $42.6 \pm 5.90$  & $5.28 \pm 3.70$ & $5100$ \\
\bottomrule
\end{tabular}
\end{table}

\section{Conclusions} \label{sec:conclusion}
We have presented a novel Bayesian optimization framework tailored for optimization under uncertainty
with a focus on tuning a scale- or precision-type parameter in stochastic models. 
The method leveraged an inexpensive statistical surrogate to evaluate expectations analytically and enable a closed-form global solution to the optimization of the acquisition function.

Through static and dynamic structural examples, we demonstrated that the method consistently achieves order-of-magnitude gains in data and computational efficiency, reducing evaluation costs by about 40-fold compared with standard one-dimensional optimization and 5 to 10-fold compared to GP-based BO.
In addition, a robustness study in which the assumed power-law scaling is intentionally violated shows that the method can reliably recover the optimal region, despite surrogate misspecification.
These results establish the proposed framework as a scalable and robust alternative for hyperparameter optimization in noisy, data-intensive settings.

While the present study focused on a single hyperparameter and assumed approximate power-law scaling, the framework can be extended to higher-dimensional parameter spaces and alternative surrogate structures.
Future work will investigate these extensions and explore applications to large-scale stochastic modeling tasks in engineering, physics, and machine learning.

\section*{Data Availability Statement}
Some or all data, models, or code generated or used during the study are available in a repository online in accordance with funder data retention policies (\href{https://github.com/UQUH/SO-BO-scale}{https://github.com/UQUH/SO-BO-scale}).

\section*{Acknowledgments}
This work was funded by the University of Houston through the SEED program no. 000189862.

\appendix
\bibliographystyle{ascelike-new}
\bibliography{HpTBO}

\end{document}